\documentclass[manuscript,screen]{acmart}
\usepackage{multirow}
\usepackage{threeparttable}
\AtBeginDocument{%
  \providecommand\BibTeX{{%
    \normalfont B\kern-0.5em{\scshape i\kern-0.25em b}\kern-0.8em\TeX}}}

\setcopyright{acmcopyright}
\copyrightyear{2023}
\acmYear{2023}
\acmDOI{XXXXXXX.XXXXXXX}

\acmPrice{15.00}
\acmISBN{978-1-4503-XXXX-X/18/06}

\begin{document}

\title{Text Classification using Graph Convolutional Networks: A Comprehensive Survey}

\author{Syed Mustafa Haider Rizvi}
\affiliation{%
  \institution{Department of Computer Science, Information Technology University}
  \city{Lahore}
  \country{Pakistan}}
\email{msds21010@itu.edu.pk}
\orcid{0009-0006-5797-8642}

\author{Ramsha Imran}
\affiliation{%
  \institution{Department of Computer Science, Information Technology University}
  \city{Lahore}
  \country{Pakistan}}
\email{phdcs22003@itu.edu.pk}
\orcid{0009-0008-9950-7253}

\author{Arif Mahmood}
\affiliation{%
  \institution{Department of Computer Science, Information Technology University}
  \city{Lahore}
  \country{Pakistan}}
\email{arif.mahmood@itu.edu.pk}
\orcid{0000-0001-5986-9876}

\renewcommand{\shortauthors}{Rizvi et al.}

\begin{abstract}
Text classification is a quintessential and practical problem in natural language processing with applications in diverse domains such as sentiment analysis, fake news detection, medical diagnosis, and document classification. A sizable body of recent works exists where researchers have studied and tackled text classification from different angles with varying degrees of success. Graph convolution network (GCN)-based approaches have gained a lot of traction in this domain over the last decade with many implementations achieving state-of-the-art performance in more recent literature and thus, warranting the need for an updated survey. This work aims to summarize and categorize various GCN-based Text Classification approaches with regard to the architecture and mode of supervision. It identifies their strengths and limitations and compares their performance on various benchmark datasets. We also discuss future research directions and the challenges that exist in this domain.
\end{abstract}



\received{20 July 2023}
\received[revised]{20 June 2024}
\received[accepted]{20 September 2024}

\maketitle

\section{Introduction}
The need for automatic text classification has been felt since the advent of digital documents. However, interest in this domain and the need to develop more efficient and robust techniques has proliferated in recent times as we as a society continue to disseminate and consume exponentially growing quantities of textual information in the form of emails, blogs, and messages on social media. Moreover, vital government and military correspondence as well as commercial documentation is also often shared and preserved in the form of electronic text documents.

In machine learning, text classification is a task that entails the categorization of text, primarily documents, sentences, or phrases, into one or more predefined categories, themes, genres, and sentiments \cite{c1}. Formally, if $D=\{d_i\}_{i=1}^m$ is the  set of  documents (or texts) and $C=\{c_j\}_{j=1}^n$ be the set of categories, then the objective of text classification is to assign a document $d_i \in D$ to its most likely category $c_j \in C$ \cite{c31, c32}.

\subsection{Text Classification by Level of Abstraction}

\begin{figure}[t]
\centerline{\includegraphics[width=0.5\textwidth]{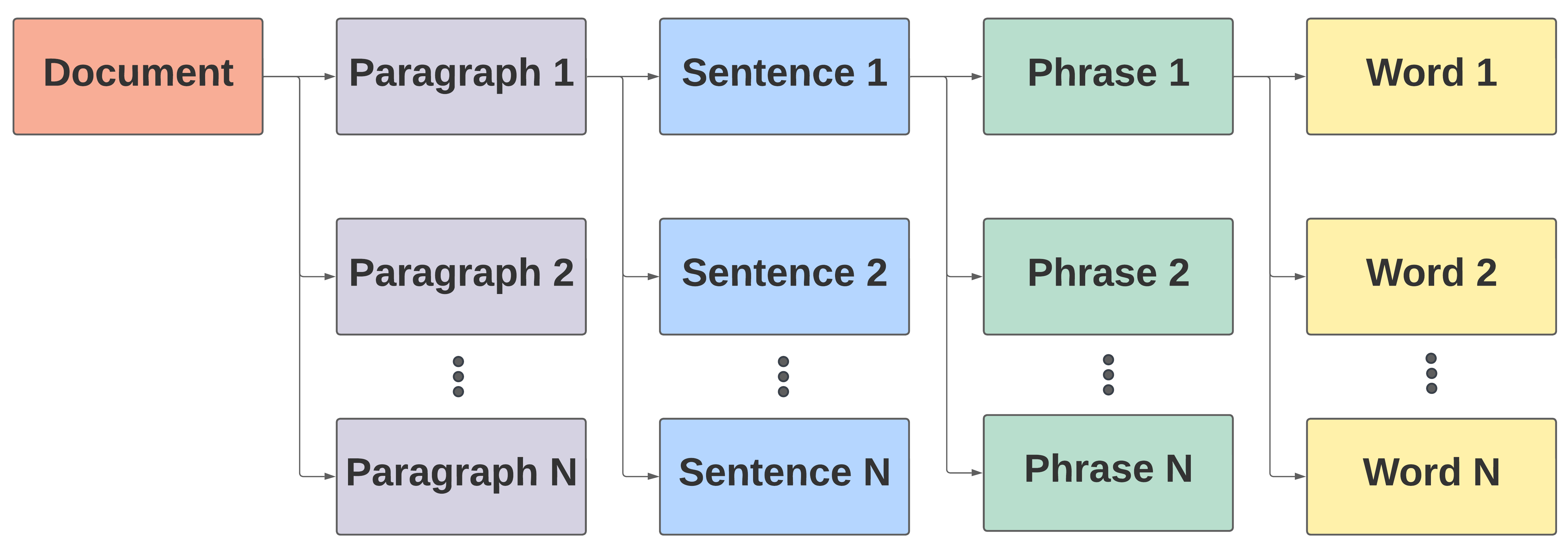}}
\caption{Different levels of document hierarchy in text classification tasks.}
\label{fig1}
\end{figure}

In general, text classification can be applied at multiple levels of document hierarchy as shown in Fig \ref{fig1}.

\begin{itemize}
\item \textbf{Word level:} This is the most common level of text classification, where the goal is to classify individual words into categories such as topics, sentiment, or named entities.
\item \textbf{Phrase level:} This involves classifying groups of words that function as a unit, such as noun phrases or verb phrases. This level of granularity is usually required in applications pertaining to sentiment classification where it would not be correct to assume that the sentiment reflected in an entire document or paragraph would remain consistent \cite{c44, c45}. A frequent challenge while attempting classification at this level is the need for extensive manual annotation for each phrase and therefore, researchers have resorted to semi-supervised methods \cite{c48}.
\item \textbf{Sentence level:} At a higher level of abstraction, sentence classification seeks to predict labels for each complete sentence \cite{c49, c50, c51}. Similar to phrase-level classification, this also enables us to derive a more meaningful description of a document by inferring distinct ideas and arguments contained within sentences instead of considering the entire text in broad strokes. As an example, this can be useful in identifying not just different speakers in a written dialogue but also changes in their attitudes and emotions.
\item \textbf{Paragraph level:} Paragraphs or subsections  of a document are classified in to a set of categories. A paragraph can have one or more sentences and can even span an entire document.  Learning distributed vector
representations for pieces of texts of variable length is an important research direction \cite{c52}.
\item \textbf{Document level:} A significant bulk of text classification  has been dedicated to document classification. At this level, the text classification algorithm predicts the classes of the entire document, i.e., the topic of an article or the genre of a book. In recent times, neural network-based approaches \cite{c53, c17} have dominated this domain.  Researchers have also attempted to exploit BERT \cite{c54} for this purpose despite its prohibitively large number of parameters and computational overhead.
\end{itemize}

\subsection{Applications of Text Classification}
Text classification has broad applications in Natural Language Processing (NLP). The most frequent examples include:
\begin{itemize}

\item \textbf{Sentiment Analysis:} Sentiment analysis classifies text as positive, negative, or neutral. It's widely used in social media monitoring, customer feedback analysis, and product review categorization. Methods like Decision Trees, Naive Bayes, and Maximum Entropy have shown high performance \cite{c113}. Moreover, deep learning techniques using CNNs \cite{c114}, RNNs \cite{c115}, and their combinations \cite{c116, c117} have also achieved promising results.

\item \textbf{Content Moderation:} Automated text classification is used to detect and flag inappropriate content on social media and online communities, including hate speech \cite{c118, c119}, offensive language \cite{c120}, adult content \cite{c121}, and privacy protection \cite{liang2024graph}, ensuring a safer user experience.

\item \textbf{Spam Filtering:} Text classification helps prevent unwanted communications from reaching a user's inbox by classifying messages as spam or not. Naive Bayes, along with SVM \cite{c123}, Decision Trees \cite{c124}, and Random Forests \cite{c125}, have proven effective in spam filtering.

\item \textbf{News Categorization:} ML automates the categorization of news articles into topics like sports, politics, and entertainment. SVM-based methods have been  used for both pre-defined \cite{c126, c127} and user-defined categories. 
\end{itemize}

\subsection{The Use of Machine Learning for Text Classification}
Over the years, a wide variety of machine learning techniques, ranging from classic techniques to deep neural networks have been leveraged for the task of text classification.
The more traditional approaches include  Naïve Bayes \cite{c4, c36}, Support Vector Machines (SVM) \cite{c37} and Decision Trees \cite{c9} as well as various ensembles and combinations of these approaches \cite{c10}.
More recent works indicate a notable transition to different neural network architectures such as Recurrent Neural Networks (RNN) \cite{c11}, and  Long short-term memory (LSTM) \cite{c12, c43}. RNNs can assign more weightage to previous data points in a sequence, which makes them suitable for understanding semantics and context while performing text classification. However, they tend to suffer from vanishing and exploding gradients, which makes it difficult for them to preserve long-term dependencies. An LSTM is a special kind of RNN that is not vulnerable to these problems and better suited for such scenarios \cite{c42}.

Convolutional Neural Networks (CNN) \cite{c41}, although originally developed for image processing, have proven to be useful for text classification as well \cite{c13, c40}. In a typical CNN, the input undergoes convolution with learnable feature maps that can be layered to offer multiple filters on the input. To decrease computational load, CNNs use pooling to reduce the output size from one layer to the next in the network. The final layers in a CNN are typically fully connected.

More recently, transformer models have demonstrated great prowess in handling a wide range of NLP tasks. Transformers \cite{c56} feature an encoder-decoder structure that relies on attention to generate an output. The encoder first maps an input sequence to a sequence of continuous representations. The decoder then takes these along with its own output at the previous time step to generate an output. Some of the more commonly used transformer-based models for text classification include BERT \cite{c14}, XLNet \cite{c57}, and RoBERTa \cite{c58}.  

Bidirectional Encoder Representations from Transformers (BERT), in particular, has achieved amazing results in many NLP tasks and text classification is no exception {\cite{zeng2024multi}}. BERT consists of multiple encoder blocks from the transformer stacked atop one another. Unlike many other transformer models, BERT is able to incorporate context on both sides of a word to gain better results. Since BERT is pre-trained on general tasks, it does not require huge amounts of additional data to fine-tune it for a particular target task. Graph Neural Networks (GNN) operate on graphs and are able to model global information in corpora. There exist several variations of GNNs, namely Graph Convolutional Networks (GCN) \cite{c16}, Graph Attention Networks (GAT) \cite{c158, c165}, Graph Auto-encoders \cite{c193}, Gated GNNs \cite{c159}, and GraphSAGE \cite{c112}. Besides these GNN architectures, there are also Graph Transformers \cite{c160, c161} that embed a graph structure into the transformer architecture \cite{c56}, enabling learning from the entire graph rather than just the local neighborhood. 

Among the various deep learning models that operate on graph data, GCNs, in particular, have shown great promise with regard to text classification \cite{c17}. One of their key strengths is their ability to factor in the global information between words and concepts by performing convolution operations on neighboring nodes in a graph to aggregate information from a node's neighbors and update that node's representation. 

While there exist several other surveys on the topic of general machine learning and deep learning text classification techniques \cite{c31, c32, c33, c34}, as well as those pertaining to graph convolutional networks and graph neural networks \cite{c29, c30, c35}, this article, to the best of our knowledge, is the first attempt at extensively and exclusively reviewing various state-of-the-art methodologies using GCNs for text classification. Significant strides have been made in this area of research in the last few years and we have therefore placed particular emphasis on the more recent GCN-based text classification methods, their conception and various strengths and weaknesses. 

\subsection{Inductive vs. Transductive Learning}
Text classification can be performed using two types of learning mechanisms:
\begin{itemize}
\item \textbf{Inductive Learning:} This mode of learning takes place in two phases, namely the training phase and the inference phase. In the training phase, we use a training set to build a machine learning model, while during inference, we generalize this model to a separate, previously unseen test set to predict its labels.

\item \textbf{Transductive Learning:} In transductive learning, both training and inference occur simultaneously. In other words, both training and test data are available to us at the same time and we make use of patterns present in both sets and labels of the training set to infer the unknown labels of the test set.
\end{itemize}

Generally, transductive learning is more computationally expensive as it requires the algorithm to rerun to infer the labels of any new datapoints whereas in inductive learning, we build a generalized predictive model beforehand that does not require retraining for inference.

\subsection{Key Findings}
{From our comparative analysis of various GCN-based text classification approaches, several key findings have emerged:
\begin{itemize}
    \item Initially, the focus was predominantly on supervised learning approaches. Early methods such as TextGCN \cite{c17} showcased remarkable performance and underscored the potential of GCNs in text classification tasks. These approaches laid the groundwork for subsequent innovations by demonstrating how graph-based representations could capture semantic relationships within text data.
    \item In terms of architecture, earlier innovations leveraged optimization-centric methods and multigraph approaches to enhance graph representations whereas more recent architectures integrate GCNs with advanced models such as BERT and other LLMs to further enhance text classification performance.
    \item As the field progressed, there was a noticeable shift towards semi-supervised methods, motivated by the need to leverage large amounts of unlabeled data, which is often more readily available than labeled datasets. 
    \item More recently, the research focus has shifted towards self-supervised learning, reflecting a broader trend in machine learning towards minimizing the dependency on labeled data.
    \item Overall, the performance on benchmark datasets has improved significantly as models have evolved from basic GCNs to more complex and hybrid architectures over the years.
\end{itemize}
}

\subsection{Criteria for Selection of Approaches and Datasets}
{In this survey, we adopted a rigorous criteria to evaluate and compare different GCN-based approaches for text classification. Specifically, we focused on seminal works as well as recent papers from the last five years that are well-cited and published in reputable venues. This approach allowed us to incorporate both historical context and the latest advancements in the field.} {Datasets commonly used across multiple studies are included in this review to ensure consistent comparisons  and to yield meaningful insights regarding their relative strengths and limitations.}

\subsection{Main Contributions of This Work}
{The key contributions of our work are outlined as follows:
\begin{itemize}
\item While there are several reviews on GCNs, 
this survey uniquely provides an exhaustive and categorical breakdown of GCN-based text classification techniques, focusing on the architecture and mode of supervision.
\item Based on supervision, we categorize existing methodologies into  supervised, semi-supervised, self-supervised and weakly supervised techniques.
\item Based on architecture, we categorize these as fundamental techniques and  GCN integration with generative models. In the latter we discuss combination of GCN with RNNs, LSTMs, BERT and LLMs, offering a structured understanding of the field.
\item Detailed analysis and comparison of various methods based on their performance on popular, benchmark datasets are provided to highlight their strengths and limitations.
\item Future research directions and challenges in GCN-based text classification are discussed to guide further advancements in the field.
\end{itemize}}

{The remainder of this survey paper is organized as follows. Section 2 provides a detailed review of the existing surveys. Section 3 provides insight into various textual embeddings. Section 4 provides an overview of the GCN architectures for text classification. Section 5 categorizes GCN architectures based on integration with generative models. Section 6 categorizes existing GCN approaches as supervised, semi-supervised, self-supervised, and weakly supervised. Section 7 covers the performance comparison, datasets, metrics and analysis of results. 
Finally, we conclude this paper in Section 8 and also discuss future research directions.}

\section{Review of Existing Surveys}
In recent years, many surveys are published for various text classification techniques in general as well as those particularly focusing on the use of graph networks for this objective. These works through effective comparisons of contemporary approaches, have not only assisted other researchers to catch up with recent developments in this domain but also identified shortcomings and potential areas for future research. 
There exist several renowned works that aim to underscore the most quintessential developments with regards to text classification techniques. Some works cover a wide range of classic ML techniques and several DL approaches for text classification \cite{c31}, \cite{c32}, \cite{c33}, and \cite{c69}. Altınel et al. discuss a wide gamut of text classification techniques and break them down into different approaches based on domain knowledge, corpus analysis, deep learning, character sequence enhancement, and linguistic enrichment \cite{c70}. However,  these works have not covered graph neural network-based techniques. Nevertheless, these surveys provide a foundational understanding of text classification and its evolution, and the key concepts they discuss still hold true.

In recent times, there has been a general transition to deep learning-based approaches for text classification as they have been found to better preserve the complex semantic relationships between words and documents as opposed to conventional ML techniques that use handcrafted features. 
Minaee et al. \cite{c34} reviewed many deep learning models for text classification including those based on recurrent neural networks, convolutional neural networks, attention-based mechanisms, and graph neural networks. While it is exhaustive with respect to its breadth, it leaves more to be desired in terms of depth. Contrarily, Pham et al. \cite{c72}, offer a comprehensive look at six  GNN-based architectures for text classification published from 2019 to 2021, and highlights their potential as well as associated limitations and challenges.

Another category of surveys includes those with a primary focus on graph neural networks instead of text classification \cite{c30}, \cite{c35}, and  \cite{c75}. These papers provide a holistic overview of developments pertaining to graph neural networks  but merely glance over their use in text classification as it was not the primary focus of these studies. Han et al. \cite{c156} focus on analyzing the graph construction and learning mechanisms for text classification, rather than comparing the performance with other baselines. Also Zhang et al. \cite{c29} and Ren et al. \cite{c77} delve much deeper and focus exclusively on GCNs and dissect them by the type of convolution operations they use as well as their areas of application, among which text classification is also discussed albeit only briefly and as an introduction.
\begin{figure*}[t]
\centerline{\includegraphics[width=0.9\textwidth]{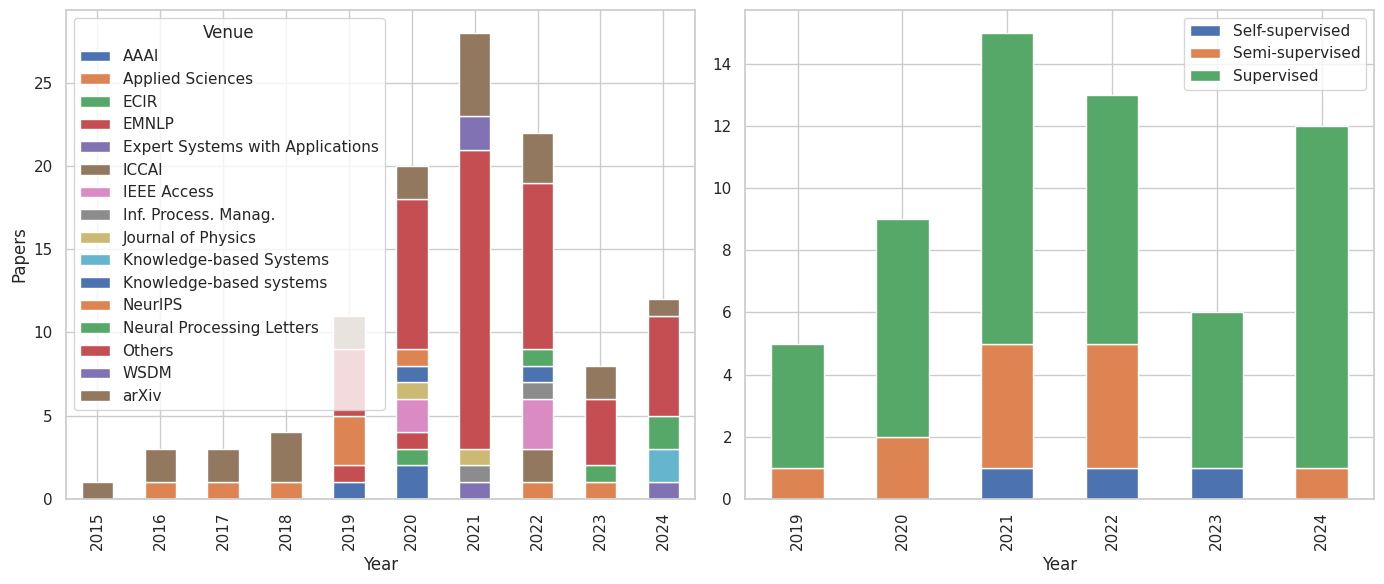}}
\caption{Statistics of Graph-based learning methods in recent years. From left to right, we show a) the overall distribution of venues in which they are published, and
b) the categorical breakdown of Graph Convolutional Network-based text classification approaches into supervised, semi-supervised, and self-supervised}
\label{fig2}
\end{figure*}

In comparison, this article provides a comprehensive and categorical review of a wide range of quintessential GCN-based approaches for text classification, along with their associated fundamentals and prospective uses. It primarily focuses on more recent state-of-the-art architectures that have yielded unprecedented results on benchmark datasets and traces their origin and evolution to earlier seminal GCN methodologies and further back to conventional machine learning and deep learning techniques. Fig \ref{fig2} shows the distributions of research articles across different publishing venues as well as across different categories of algorithms based on supervision. It gives the reader a quick overview of the field regarding text classification using GCN-based approaches. 

\section{Overview of Textual Embeddings}
Text embedding techniques generate high-dimensional vector representations for text, capturing semantic and syntactic relationships between words. These embeddings are crucial for tasks like text classification, sentiment analysis, and language translation, potentially improving performance and reducing training time. The choice of embeddings can significantly impact task success, as different embeddings capture various text aspects \cite{c102, c103, c104, c105}.

\begin{itemize}
\item \textbf{{Term Frequency-Inverse Document Frequency (TF-IDF)}:} Maps words into a feature space based on their frequency in a document and rarity in the corpus. Since the vocabulary may contain millions of words, such models are hard to scale. Moreover, unlike contextualized embeddings \cite{c109}, TF-IDF is unable to account for the similarity between the words in a document since each word is encoded in isolation.

\item \textbf{{Word2Vec}:}
Word2Vec \cite{c106} represents words as vectors learned from context by predicting a target word based on the context words around it (Continuous Bag-of-Words) or by predicting the context words given the middle word (Skip-gram). Word2Vec can capture important semantic and syntactic relationships between words while being computationally efficient compared to other prediction-based embedding techniques.

\item \textbf{{Doc2Vec}:}
Doc2Vec \cite{c52} extends Word2Vec to learn vector representations for entire documents or paragraphs, handling variable-length texts with fixed-length vectors. Analogous to Word2Vec’s Continuous Bag-of-Words and Skip-gram, Doc2Vec has the distributed Bag-of-Words model and the distributed memory model, respectively, to learn distributed representations of documents.

\item \textbf{{GloVe}:}
Global Vectors (GloVe) \cite{c108} generates word vectors by factorizing a global word-word co-occurrence matrix, capturing semantic and syntactic relationships between words and understanding word meaning across larger contexts through aggregated statistics.

\item \textbf{{Graph Embeddings}:}
Graph embeddings represent nodes and edges in a graph as high-dimensional vectors. While studies have demonstrated the ability of techniques like Word2Vec and GloVe to capture global connections between words in a language, these connections are still rather limited. Graph embeddings go beyond this and capture dependencies over a much longer range, leveraging the structural and semantic properties of the graph.
Random walk-based methods such as DeepWalk \cite{c110} and Node2Vec \cite{c111} generate embeddings by performing random walks on the graph and using the sequence of nodes visited as the context for the node. Deep learning-based techniques such as GCNs \cite{c16} and GraphSAGE \cite{c112} generate embeddings using neural networks to aggregate information from the node's local neighborhood.

\item \textbf{{LLM-Based Embeddings}:} LLMs typically utilize embedding models that are integral to their architecture for representing words, tokens, or sub-words. These embeddings are learned jointly with the model during pre-training. In transformers such as GPT, tokens are initially input using basic embeddings \cite{c56}, which are then processed through multiple transformer layers to get contextual embeddings. 
WordPiece tokenizer breaks words into smaller units called sub-word tokens \cite{c195,c200} and  each sub-word token is assigned an embedding vector. Similarly  SentencePiece \cite{c196}  tokenization is used in models like T5 (Text-To-Text Transfer Transformer) \cite{c92}. It segments text into smaller units and assigns embeddings to these units based on their context.
Positional embeddings are added to word embeddings to encode the position or order of tokens in a sequence. They help the model understand the sequential structure of input sequence. Models like ELMo \cite{c198} and GPT \cite{gpt35} use contextual embeddings, where each token's embedding depends not only on the token itself but also on the entire input sentence. These embeddings capture richer semantic and syntactic information.

\item \textbf{{BERT Embeddings}:}
Based on the transformer architecture, Bidirectional Encoder Representations from Transformers (BERT) \cite{c109} is an advanced pre-trained embedding model that generates unique vector representations that account for the context of a word within a sentence. A deeply bidirectional system, BERT can understand the context of words to the left and to the right of a given word in a sentence. This affords it more power than earlier models, which were unidirectional.
BERT embeddings are based on the internal representations of the model learned during pre-training on a large corpus of text. The pre-training involves training the model on two tasks: masked language modeling (predict masked words in a sentence) and next sentence prediction (predict whether two sentences are adjacent).
The embeddings generated by BERT capture both the meaning and syntax of words along with the relationship between words in a sentence. This makes them particularly useful for tasks that require understanding the context of words in a sentence, such as text classification and named entity recognition.
\end{itemize}

\section{Overview of GCN Architectures for Text Classification}
In this section, we provide an overview  of Graph Convolutional Networks (GCNs) based on the findings of Kipf and Welling \cite{c16}, and some fundamental concepts. Additionally, we will overview some foundational GCN architectures that have significantly contributed to the  advancement of GCNs for text classification.

\subsection{GCN Preliminaries}
Let $G = (V, E)$ be a graph where $V$ and $E$ represent sets of nodes and edges respectively. Each node $v_{i}$ in the graph has a corresponding m-dimensional feature vector $x_{i} \epsilon R^{n}$. If we have $n$ nodes, this translates to a feature matrix $X \epsilon R^{nxm}$ that serves as the input to the first GCN layer. 
In supervised classification, the labels of only a subset of nodes is known and the objective is to predict the remaining unknown labels. 


Adjacency matrix $A$, the Degree matrix $D$, and the normalized adjacency matrix $\hat{A}$ are often used as Graph Shift Operators (GSO). $A$ is a sparse matrix with non-zero entities with all the diagonal elements set to one since each node is assumed to be connected to itself. For its remaining elements, we set $A_{ij} = w_{ij}$, for all ($i$, $j$) where $w_{ij}$ is the edge weight between nodes $i$ and $j$. $D$ is a diagonal matrix with degrees as diagonal entries, i.e., $D_{ii} = d_i$. The normalized adjacency matrix is given by $\hat{A} = D^{-1/2}AD^{-1/2}$.

\subsection{Graph Convolutional Networks}
A GCN is a multilayer neural network that can be applied directly to graphs, yielding vectors of nodes based on their connections with other nodes and overall graph topography. For a single layer GCN, a graph convolution can be represented as: $H_1 = \rho(\hat{A}XW_0)$, where $W_0$ is a learnable weight matrix for the first layer and $\rho$ is an activation function like ReLU. Such a GCN can only update a node's representation based on information captured from its immediate neighbors. To leverage information from distant neighbors, multiple GCN layers can be stacked as given:
$H_{k+1} = \rho(\hat{A} H_{k} W_{k}).$
\begin{figure}[t]
\centerline{\includegraphics[width=0.8\textwidth]{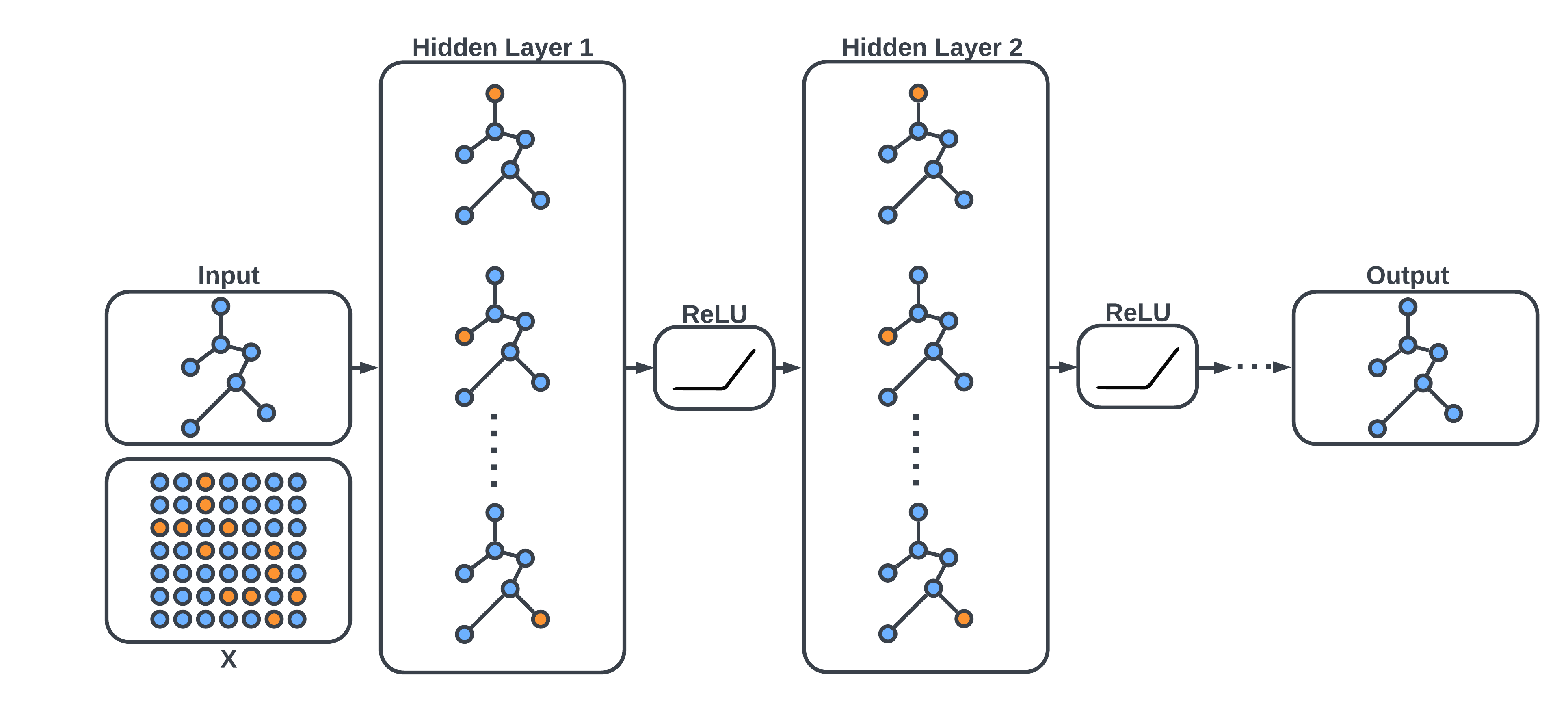}}
\caption{A typical GCN with two Graph Convolutional layers, each with ReLU activation function. The nodes with known labels are shown in orange while test nodes are shown in blue.}
\label{fig3}
\end{figure}
A k-layer GCN as depicted in Figure \ref{fig3} can thus allow message passing among nodes that are at maximum k hops away, which tends to have a similar effect as an increased receptive field in deeper CNNs. Successive convolutions smoothen the resulting hidden representations locally along the edges of the graph, resulting in similar predictions among closely connected nodes.  Typically, 2-layer GCNs have been used in literature as a higher number of layers yields diminishing returns and even worse performance due to over-smoothing. Predictions are made using a softmax classifier at the output whereas cross-entropy error over all labeled nodes is used as the loss function.

\subsection{Fundamental GCN Approaches for Text Classification}
Graph-based approaches for text classification have long existed \cite{c157}. Defferrard et al. \cite{c15} were among the first to extend and demonstrate the efficacy of Spectral Graph Convolutional Neural Networks for this task. They framed their problem as a semi-supervised node classification problem where labels were only available for a small subset of nodes and based it on the assumption that connected nodes likely share the same label. They leveraged a global graph with Word2Vec embeddings and fast K-localized convolutions were applied. 

Kipf and Welling \cite{c16} put forward a solution based on spectral GCNs featuring fast localized convolutions.  They also considered transductive node classification in notably larger networks. They demonstrated that with a layer-wise propagation rule along with fewer parameters and operations, both scalability and classification performance can be improved in large-scale networks. This method proved to be a seminal work in this domain as it formalized GCNs as we know them and laid the foundation for all GCN methods to come. Besides being extensively applied to word-document graphs for text classification in subsequent literature, several works have also built upon this architecture for semi-supervised node classification. Specifically by introducing a discriminative hierarchical convolutional mechanism \cite{c189}, incorporation of high-order proximities using quantum information theory \cite{c190}, integrating with a neural topic model \cite{c191}, and including more robust privacy measures \cite{c192}.
\begin{figure}[t]
\centerline{\includegraphics[width=0.8\textwidth]{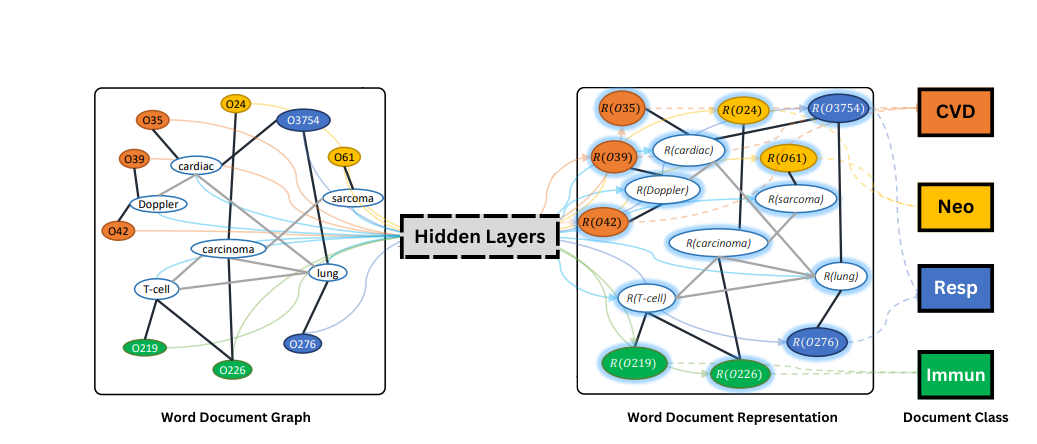}}
\caption{Illustration of Text GCN architecture by Yao et al. \cite{c17}. Nodes starting with 'O' are document nodes, all others are word nodes. Bold black edges link document-word nodes, while thin gray edges link word-word nodes. $R(x)$ represents the resulting embedding of $x$. Sample data is from the Ohsumed database and different colors depict different classes (CVD, Neo, Resp, Immun).}
\label{fig4}
\end{figure}
Yao et al. \cite{c17} built upon the work of \cite{c15} and \cite{c16} by leveraging a singular heterogeneous text graph for the entire corpus and further extended them through the inclusion of document nodes and the representation of relationships among word and document nodes using suitable metrics such as Pointwise mutual information for word-to-word relations and TF-IDF for word-to-document relations as shown in Fig. \ref{fig4}. They then modeled this graph with GCN \cite{c16} so it may automatically learn embeddings of both words and documents in conjunction, supervised by a small subset of labeled documents. The text graph construction method and application of a two-layer GCN presented in this paper has influenced many later works \cite{c18, c19, c20, c173, c129, c23, c171, c170, c169, c172, c25, c26, c166, c168, c167, c164, c131, c97, c98, c24, c94, c95, c96, c174, c101, c99, c100} as discussed in the following sections.

\subsection{Integration of GCN with CNN}
Some researchers have sought to use Convolutional Neural Networks (CNNs) with GCN, owing to the former’s ability to capture local contextual information effectively. Zeng et al. \cite{c187} proposed a GCN-CNN boosting ensemble in which a CNN learned from examples misclassified by a GCN to improve performance. Similarly, in addition to presenting a hybrid GCN-Bi-LSTM model, Yang et al. \cite{c182} also introduced variant models (ServeNet, C-LSTM) that used 1-D or 2-D CNN layers in addition to the Bi-LSTM layers to better capture local information.
\begin{table}[t]
\centering
\resizebox{0.9\textwidth}{!}{
\begin{tabular}{|p{4cm}|p{8cm}|p{5.5cm}|}
\hline
\textbf{Category} & \textbf{Description} & \textbf{Methods} \\
\hline
Fundamental GCN Approaches & Seminal techniques that formalized GCNs for text classification and paved the way for further innovations. & Graph-CNN (2016) \cite{c15}, GCN (2017) \cite{c16}, TextGCN (2019) \cite{c17} \\
\hline
Integration with CNNs & Architectures combining GCNs with CNNs to enhance feature extraction by leveraging graph structures and spatial hierarchies in text data. & TextGCN C-LSTM, TextGCN ServeNet (2021) \cite{c182}, GCN-CNN (2022) \cite{c187} \\
\hline
Integration with RNNs, LSTMs, GRUs & Architectures combining GCNs with RNNs and variants like LSTM and GRU to capture sequential dependencies and graph structures. & IGCN (2020) \cite{c180}, GCN-LSTM (2020) \cite{c181}, GL-GCN (2021) \cite{c183}, TextGCN Bi-LSTM (2021) \cite{c182}, BiGRU+GCN (2022) \cite{c184} \\
\hline
Integration with Transformer & Methods combining GCNs with Transformer models to leverage self-attention mechanisms and capture long-range dependencies in text. & GTG (2023) \cite{c188}, TLC-XML (2024) \cite{zhao2024tlc} \\
\hline
Integration with BERT & Hybrid models combining GCNs with BERT to capture both graph and contextual information for enhanced performance. & VGCN-BERT (2020) \cite{c21}, GC-GCN-BERT (2021) \cite{c176}, MGCN (2021) \cite{c175}, R-GCN (2022) \cite{c178}, BERT-GCN + MA (2022) \cite{c177}, BERT-GCN, RoBERTa-GCN (2022) \cite{c22}, HINT-G (2023) \cite{c179} \\
\hline
Integration with BERT+LSTMs & Models combining GCNs, BERT, and LSTMs to exploit graph structures, contextual embeddings, and sequential dependencies. & WordBERT-BiLSTM-SGCN (2021) \cite{c185}, IMGCN (2022) \cite{c153} \\
\hline
Integration with LLMs & Models that integrate GCNs with large language models to leverage extensive pre-trained knowledge for improved text processing. & Clip-GCN (2024) \cite{zhou2024clip}, GCN+GPT (2024) \cite{chen2024exploring}, GCN+Llama2-13B (2024) \cite{li2024ontology}, Graph Aware Convolution + ChatGPT (2024) \cite{du2024large} \\
\hline
\end{tabular}
}
\caption{Architecture-based Categorization of GCN Approaches for text classification}
\label{tab1}
\end{table}
\begin{figure*}[b]
\centerline{\includegraphics[width=0.9\textwidth]{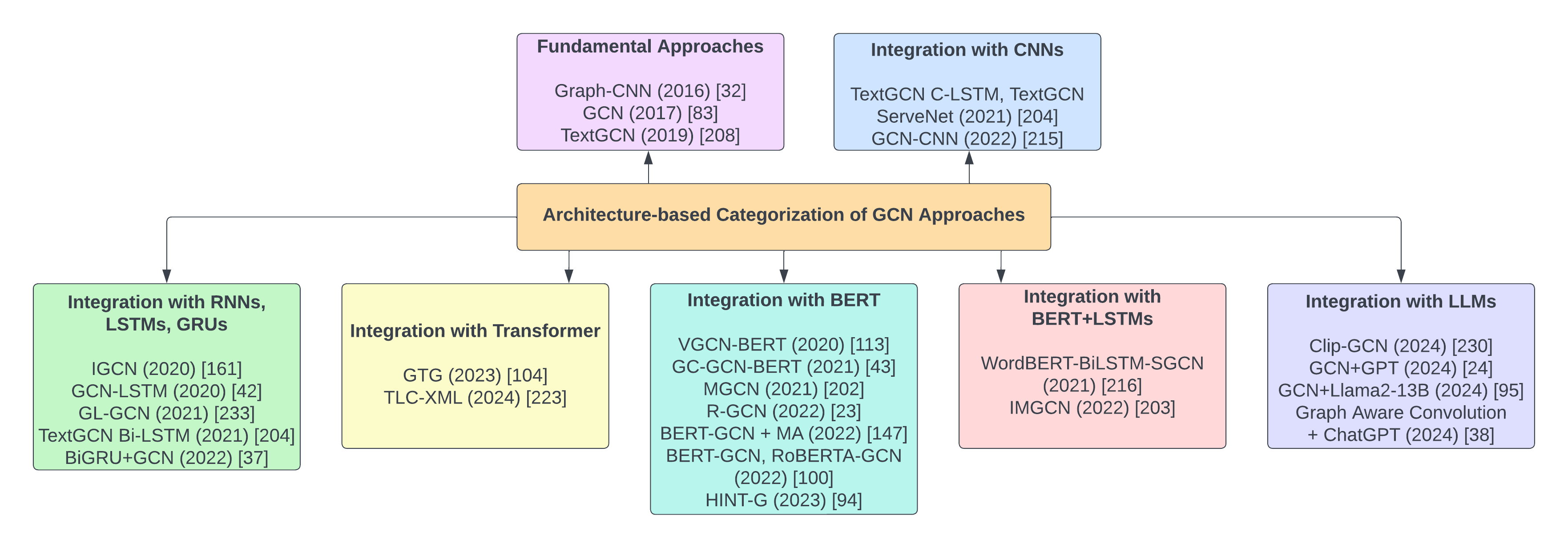}}
\caption{Architecture-based Categorization of GCN Approaches for text classification. A further division of approaches based on integration with generative models is also shown.}
\label{fig5}
\end{figure*}
\section{Integration of GCN  with Generative Models}
{
There is a notable body of work dedicated to improving the text classification performance of GCNs by augmenting them with various state-of-the-art generative models as shown in Table \ref{tab1} and Fig. \ref{fig5}. The most common approaches involve integrating GCNs with recurrent neural networks (RNNs), LSTMs, and GRUs. Additionally, GCNs have been combined with large language models (LLMs), BERT and its many variants, showcasing the versatility and enhanced performance achieved through these hybrid architectures.}

\subsection{GCN Integration with RNNs, LSTMs, and GRUs}
GCNs have been used in tandem with recurrent neural networks (RNN) and their variants, such as Long Short-Term Memory (LSTM) and Gated Recurrent Unit (GRU). The primary goal here is to enable the overall architecture to effectively capture long-range and short-range contextual dependencies. Tang et al. \cite{c180} combined textual and part-of-speech features obtained using Bi-LSTM with an adjacency matrix capturing the dependency relationship to address the problems of contextual dependency and lexical polysemy in GCNs.
Gao et al. \cite{c181} also proposed a GCN and LSTM hybrid structure that combined outputs of each GCN layer with an embedding generated by passing them through an LSTM. Meanwhile, Yang et al. \cite{c182} leveraged a weighted combination of evaluation values from GCN and Bi-LSTM classifiers to improve text classification performance.
Similarly, Zhu et al. \cite{c183} utilized a Bi-LSTM to capture the local structure information of a sentence and a text graph and GCN to model the global dependency information between words. Both global and local dependency structure signals were then fused using an attention mechanism and used to guide the training process.

In terms of performance, GRUs are generally faster and less complex than LSTMs because they have fewer gates. Thus, Dong et al. \cite{c184} opted to use a two-way GRU model in tandem with a GCN in their proposed architecture. They fed Word2Vec embeddings into a BiGRU layer, yielding a representation that captured the global contextual features and long-range dependencies within the text. This representation was then passed through a GCN layer to extract complex semantic relations and spatial feature information. The GCN output was then fed into a classifier that predicted the class label of the input text.

\subsection{GCN Integration with Transformer}
Liu et al. \cite{c188} proposed a  method that combined Transformer with GCN for improved semantic representation of documents. After the text graph underwent the first graph convolutional layer, its word nodes were fed into a transformer to capture the contextual and sequential information of the text. The graph was then augmented with the transformer’s outputs and passed through a second graph convolutional layer to ultimately yield the final classification results. TLC-XML \cite{zhao2024tlc} is a Transformer-based model for extreme multi-label classification. It employs GCNs for cluster correlation learning. 

\subsection{GCN Integration with BERT}
\begin{figure}[t]
    \centering
    \includegraphics[width=0.8\linewidth]{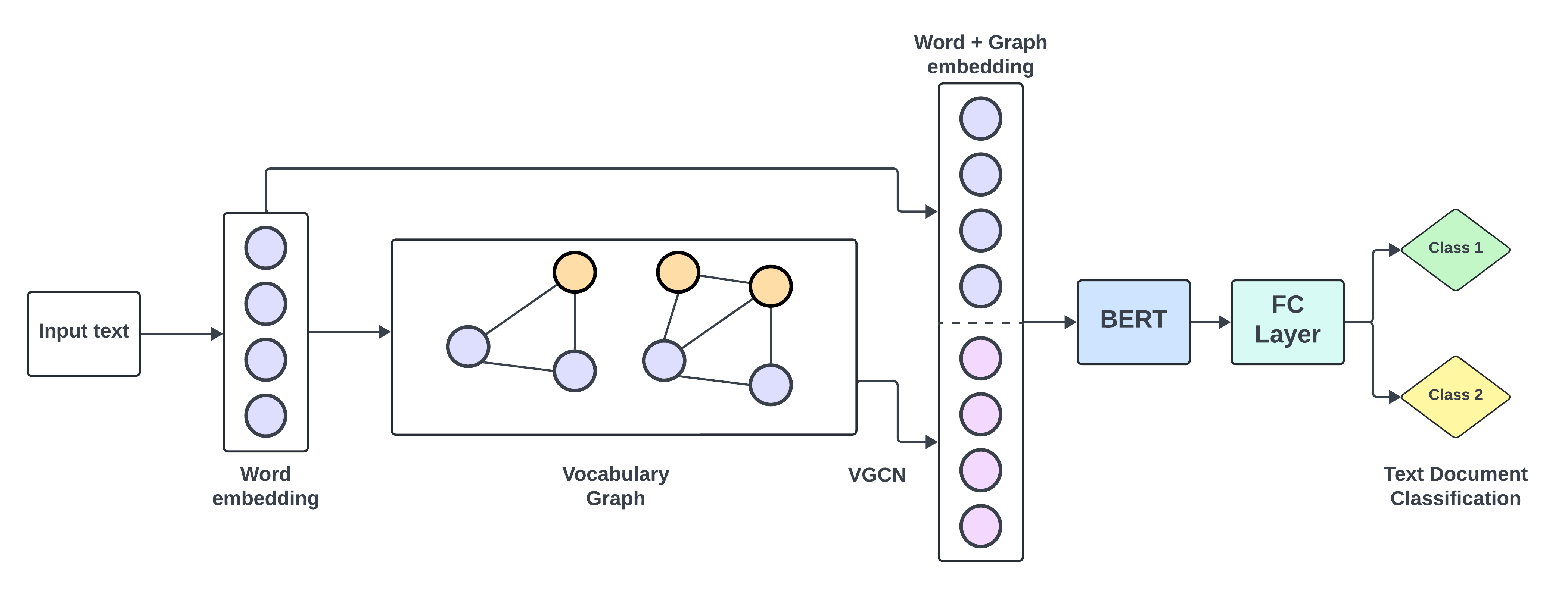}
    \caption{Illustration of VGCN-BERT, proposed by Lu et al. \cite{c21}. Input text word embeddings are combined with the relevant part of the vocabulary graph to produce a graph embedding, which is concatenated with the input sentence. The BERT model then applies self-attention to this concatenated representation, enabling interactions between word and graph embeddings. The final embedding is fed into a fully connected layer for classification.}
    \label{fig6}
\end{figure}

{
Lu et al. \cite{c21} augmented GCN's capacity to model global information about the vocabulary of a language with BERT’s ability to capture the local contextual information within a sentence or document and proposed a solution that outperformed either of its individual components as evident by their experiments on various state-of-the-art text classification datasets. They first built a GCN on the vocabulary graph based on word co-occurrence information, similar to approaches in \cite{c15, c16, c19}, and then passed the word embedding and the relevant part of the graph embedding together to a self-attention encoder in BERT. This enabled them to interact and guide each other while learning the classifier such that the resulting representation could harness local and global information when performing text classification (see Fig. \ref{fig6}).
Xue et al. \cite{c175} applied a GCN to a text graph constructed based on NPMI and WordNet to obtain a hidden state representation that could effectively capture global contextual dependencies and semantic information. An attention mechanism was then used to combine this with the local information extracted using BERT. Alternatively, Gao and Huang \cite{c176} and She et al. \cite{c177} used a gating mechanism to integrate BERT and GCN embeddings.
Chen et al. \cite{c178} proposed a relational graph convolutional network to process the semantic features obtained by BERT representation as part of a text-based accident causal classification method. The R-GCN captured immediate syntactic neighbor information of each word and assigned different weights to different types of edges. They also introduced a gate mechanism to reduce the influence of false dependency edges caused by the domain gap.
Lin et al. \cite{c22} proposed a method that first constructed a heterogeneous graph for the corpus with nodes representing either words or documents, similar to TextGCN. However, the document node embeddings were initialized with pre-trained BERT representations. Thus, by jointly training the BERT and GCN modules, this model could take advantage of both BERT’s large-scale pretraining and GCN’s message propagation mechanism, the combined effectiveness of which had not been previously explored. The resulting model was able to obtain state-of-the-art performance on a wide range of text classification datasets.
More recently, Li et al. \cite{c179} used a pre-trained BERT model to initialize document nodes in their heterogeneous graph structure that also included word and entity nodes. They generated entity nodes by mapping the entities in the text to an exogenous knowledge base. While word-word and document-word edges were modeled as in \cite{c17}, an attention mechanism was used to weigh document-entity edges and personalized PageRank to measure the semantic relatedness of entity-entity edges. An enhanced GCN with graph sampling and DropEdge techniques was then applied to mitigate the problems of neighbor explosion and noisy inputs in GCN, and the final node embeddings were used for text classification.
}

\subsection{GCN Integration with BERT+LSTMs}
Several hybrid GCN architectures have also been explored in recent research. For instance, various authors have leveraged GCN, BERT, and Bi-LSTM simultaneously to improve text classification performance. Zeyu et al. \cite{c185} used BERT to obtain word representations from long texts and fed them into a Bi-LSTM model to capture their semantic relationship. A GCN was then applied to a graph constructed using these word features as nodes and vectors similarity between them as edges.
Xue et al. \cite{c153} also used BERT embeddings as input features along with one graph based on NPMI and WordNet and the other based on dependency relationships. GCNs were trained on these graphs separately. The resulting hidden states were concatenated, fused with the input BERT features, and fed into a fusion model composed of two Bi-LSTM layers with an attention layer in between to generate the final features for classification.

\subsection{GCN Integration with LLMs}
The synergy between GCNs and the LLMs in NLP enhances performance in text classification. Zhou et al. \cite{zhou2024clip} proposed Clip-GCN multimodal fake news detection, which leveraged the CLIP \cite{radford2021learning} pre-training model to extract joint semantic features from image-text information. The model utilized adversarial neural network  to extract inter-domain invariant features and employed GCNs to capture intra-domain knowledge for detecting emergent news. Chen et al. \cite{chen2024exploring} used, gpt-3.5-turbo-0613 \cite{gpt35} as LLM, in graph machine learning for text classification. They investigated two approaches: LLMs-as-Enhancers and LLMs-as-Predictors. The former enhanced nodes' text attributes with LLMs' knowledge and generated predictions using MLP, GCN, and GAT, comparing their performance. Li et al. \cite{li2024ontology} proposed a hybrid approach that combined Natural Language Inference (NLI) and Graph Convolutional Networks (GCNs) for ontology completion. The GCN models used ConCN concept embeddings as input features and achieved strong performance, with the GCN and Llama2-13B \cite{touvron2023llama} variant. Du et al. \cite{du2024large} proposed a Graph-aware Convolutional LLM method aimed at enabling LLMs to capture high-order relations within user-item graphs using textual data. This method employed the LLM as an aggregator in graph processing to facilitate a step-by-step understanding of graph-based information. Specifically, the approach leveraged ChatGPT to enhance descriptions by systematically exploring multi-hop neighbors layer by layer, thus progressively propagating information throughout the GCN. 
\subsection{Discussion}
{
The integration of GCNs with generative models has emerged as a powerful approach for enhancing text classification performance.  Various hybrid architectures are developed, leveraging the strengths of GCNs in modeling relational structures and the advanced contextual understanding provided by models like BERT, LLMs, and LSTMs. This integration have consistently led to improvements in accuracy and robustness across diverse datasets.} {Early models combining GCNs with BERT, such as VGCN-BERT \cite{c21} and GC-GCN-BERT \cite{c176}, successfully merged global and local text features, resulting in superior performance. Subsequent models, such as R-GCN \cite{c178} and BERT-GCN \cite{c22}, further enhanced this synergy by introducing gating and attention mechanisms.} {The integration of GCNs with LLMs represents a more recent and cutting-edge development. Models like Clip-GCN \cite{zhou2024clip} and GCN+GPT \cite{chen2024exploring} have utilized the extensive pre-trained knowledge of LLMs to enrich the text attributes within GCNs, significantly boosting performance in tasks such as fake news detection.}

{Incorporating GCNs with RNNs, including LSTM and GRU variants, has also proven effective. These architectures benefit from RNNs' capability to model sequential dependencies, which complements GCNs' graph-based relational understanding. Hybrid models like GCN-LSTM \cite{c181} and BiGRU+GCN \cite{c184} have leveraged this dual capability to capture both long-range dependencies and local contextual information, resulting in improved text classification outcomes.}{Beyond these combinations, GCNs have been integrated with a variety of other architectures, including CNNs and Transformers, to further enhance their performance. Models like GCN-CNN \cite{c187} and GTG \cite{c188} benefit from CNNs' ability to capture local features and Transformers' capacity for contextual understanding, resulting in robust and accurate text classification systems.} {From early GCN-BERT combinations to the latest GCN-LLM integrations, the augmentation  of GCNs with generative models has consistently driven advancements in text classification. By leveraging the complementary strengths of GCNs and various generative models, researchers have developed increasingly sophisticated and effective text classification techniques, leading to continuous improvements in performance across a wide range of applications.}

\section{Supervision-based Categorization of GCN Approaches} \label{categories}

\begin{table}[t]
\resizebox{0.9\textwidth}{!}{
\begin{tabular}{|p{2.5cm} |p{3.2cm}|p{5.5cm}|p{5cm}|}
\hline
\textbf{Category} & \textbf{Description} & \textbf{Sub-category} & \textbf{Methods} \\ \hline
\multirow{6}{*}{Supervised} & \multirow{6}{*}{\begin{tabular}[c]{@{}l@{}}Require labeled data for\\ training and make\\ predictions accordingly.\end{tabular}} & Optimization-centric & TL-GNN (2019) \cite{c18}, SGC (2019) \cite{c19}, SSGC (2021) \cite{c20}, NMGC (2021) \cite{c173}, LDGCN (2023) \cite{c129} \\ \cline{3-4} 
 &  & Multigraph & TensorGCN (2020) \cite{c23}, SK-GCN (2020) \cite{c171}, GFN (2022) \cite{c169}, KG-GCN (2023) \cite{c172} \\ \cline{3-4} 
 &  & Inductive & TextING (2020) \cite{c25}, InducT-GCN (2022) \cite{c26} \\ \cline{3-4} 
 &  & Multilabel & HBLA (2020) \cite{c166}, LDGN (2021) \cite{c168}, GCN-BERT (2022) \cite{c167} \\ \cline{3-4}
 &  & Classification with Class Imbalance & MMCT-GCN (2023) \cite{c205}, GNN-AWB (2023) \cite{c209}, MCICIT (2024) \cite{he2024mcict} \\ \cline{3-4}
 &  & Extreme Text Classification & TLC-XML (2024) \cite{zhao2024tlc} \\ \cline{3-4}
 &  & Multilingual & CLHG (2021) \cite{c170}, MSA-GCN (2024) \cite{mercha2024heterogeneous} \\ \cline{3-4} 
 &  & Hierarchical & AMKI-HTC (2024) \cite{feng2024adaptive} \\ \hline
\multirow{8}{*}{\begin{tabular}[c]{@{}l@{}}Semi-supervised\end{tabular}} & \multirow{8}{*}{\begin{tabular}[c]{@{}l@{}}Use a small amount of\\ labeled data and a large\\ amount of unlabeled\\ data to improve\\ training.\end{tabular}} & Short Text Classification & STGCN (2020) \cite{c164}, HGAT (2021) \cite{c131}, MP-GCN (2022) \cite{c97}, ST-TextGCN (2022) \cite{c98} \\ \cline{3-4} 
 &  & Multigraph & TextGTL (2021) \cite{c24} \\ \cline{3-4} 
 &  & Zero-Shot Classification & ZS-TC (2021) \cite{c94} \\ \cline{3-4} 
 &  & Inductive & HeteGCN (2021) \cite{c95}, HDGAT (2024) \cite{lin2024heterogeneous} \\ \cline{3-4} 
 &  & Document-Document Edge Definition & ME-GCN (2022) \cite{c96} \\ \cline{3-4} 
 &  & Multi-Task Classification & MT-TextGCN (2022) \cite{c174} \\ \cline{3-4} 
 &  & Neighborhood-level Contrastive Learning & NNC-GCN (2024) \cite{xiao2024nnc} \\ \hline
\multirow{4}{*}{\begin{tabular}[c]{@{}l@{}}Self-supervised\end{tabular}} & \multirow{4}{*}{\begin{tabular}[c]{@{}l@{}}Generate supervisory\\ signals from input data\\ to train without labels.\end{tabular}} & Multimodal Representation Learning & GCNW-FL (2021) \cite{c101} \\ \cline{3-4} 
 &  & Contrastive Learning with Augmentation & CGA2-TC (2022) \cite{c99} \\ \cline{3-4} 
 &  & Pre-Trained Language Model Integration & Cont-GCN-BERT, Cont-GCN-XLNet, Cont-GCN-RoBERTa (2023) \cite{c100} \\ \hline
 \multirow{1}{*}{Weakly Supervised} & \begin{tabular}[c]{@{}l@{}}Use noisy or limited\\ labels when fully labeled\\ data is scarce or costly.\end{tabular} & Multiple Instance Learning & GNN for MIL (2019) \cite{c211} \\ \hline
\end{tabular}
}
\caption{Supervision-based Categorization of GCN Approaches for Text Classification}
\label{tab2}
\end{table}

\begin{figure*}[b]
\centerline{\includegraphics[width=0.9\textwidth]{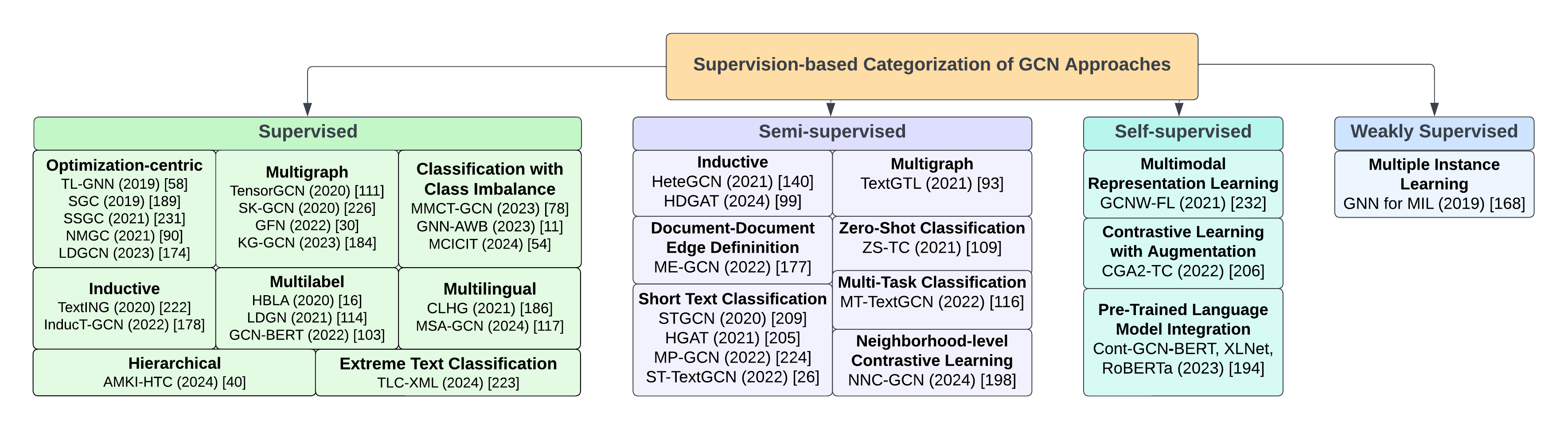}}
\caption{Supervision-based categorization of text classification approaches using GCN.}
\label{fig8}
\end{figure*}

{
In this section, we summarize various GCN-based approaches that innovated upon the foundation laid by Yao et al. \cite{c17} and categorize them by their mode of supervision, i.e., supervised, semi-supervised, self-supervised, and weakly supervised (see Fig. \ref{fig8} and Table \ref{tab2}).
}

\subsection{Supervised Text Classification}

In supervised text classification, the GCN model is trained using labeled data. More specifically, nodes of documents that form the training set make use of associated labels from a set of one or more predefined classes to train the model such that it can predict labels for unseen documents. During training, the model’s parameters are optimized using a loss function, such as Binary or Categorical Cross-Entropy, which measures the difference between the model’s predicted outputs and the true labels of the training examples.

{
The following approaches aim to build upon TextGCN in a multitude of ways, i.e., optimizing the architecture for improved efficiency, stacking multiple graphs to capture additional context, or addressing its limitations so it may effectively perform in inductive settings.}

\subsubsection{Optimization-centric Approaches}

The aforementioned methods rely on building a single corpus-level graph, for which training can prove highly cumbersome and computationally taxing as the vocabulary increases. A large vocabulary size in turn increases the number of nodes with an even greater increase in the number of edges. Consequently, matrix operations for deriving Graph shift operators and their subsequent application while applying k-localized convolutions also becomes computationally expensive. Researchers have attempted to address this issue and proposed various computationally efficient algorithms.

 Huang et al. \cite{c18} refrained from applying convolution operations to a global graph, citing memory consumption as a possible problem. They instead produced document-level graphs for each input article by connecting word nodes within them. These more local representations were then shared globally, allowing their weights to be updated through a message-passing mechanism, where a node takes in the information from neighboring nodes to update its representation, essentially preserving context. Finally, representations of all the nodes were summarized in the graph to predict the results while consuming notably less memory.

Wu et al. \cite{c19} approached the issue of increasing computational complexity from another angle. They argued that GCNs were unnecessarily complex and computationally redundant owing to origins rooted in prior deep learning approaches. Their Simple Graph Convolution approach (SGC) attempted to reduce this excess complexity by removing non-linearities and in turn, collapsing weight matrices between successive layers of the GCN to yield a linear model. This essentially reduced the entire procedure to a simple feature propagation step (applying the K-th power of the normalized adjacency matrix in a single-layer neural network) followed by a standard logistic regression layer instead of softmax as in a standard GCN. Their results demonstrated that their simplified model was more computationally efficient and scalable than its nonlinear counterparts while still achieving similar and, in some cases, superior classification performance.

Building upon this approach, Zhu and Koniusz \cite{c20} also leveraged a linear model that enabled them to keep computational costs down. However, unlike SGC, their approach, Simple Spectral Graph Convolution (SSGC) was able to keep over-smoothing in check when a larger number of graph convolutional layers were applied while also preserving the large context of each node. This was done by preventing the largest neighborhoods from over-dominating while aggregating over neighborhoods of gradually increasing sizes.

Moreover, Lei et al. \cite{c173} also addressed overfitting concerns and improved computational efficiency while demonstrating similar performance to TextGCN by proposing a weight-sharing mechanism that enabled them to use the same weight matrix for different order graph convolutions. By then fusing different neighbor features from 1-hop to k-hops using a  multi-hop neighbor information fusion mechanism, they were able to capture additional information without an increase in the number of parameters.

Wang et al. \cite{c129} introduced a new discriminative objective function that minimized the intra-class distance and maximized the inter-class distance in the resultant features of the texts while also minimizing the classification cross-entropy loss function. Thus, by jointly training on these objectives, their model was able to learn representative embeddings while utilizing the intra and inter-class manifold structures inherent to the graph.
\begin{figure}[t]
    \centering
    \includegraphics[width=\linewidth]{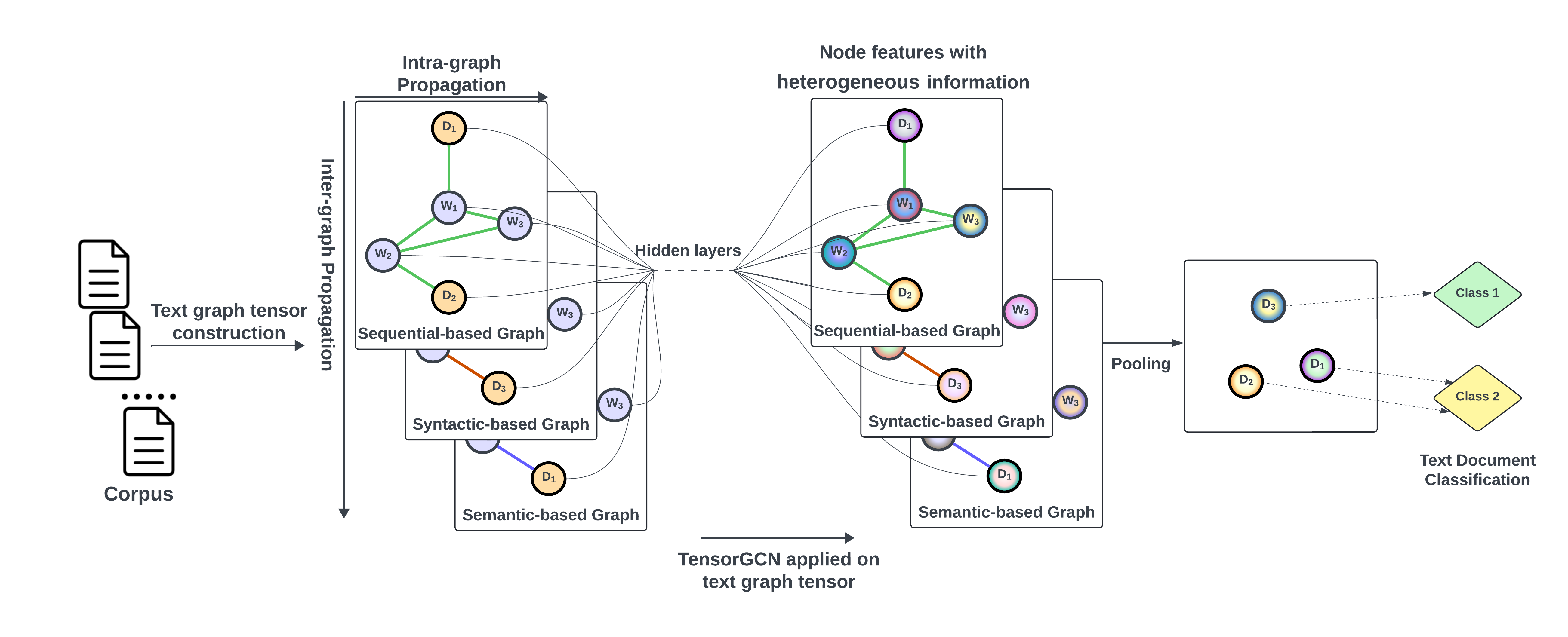}
    \caption{The TensorGCN proposed by Liu et al. \cite{c23}. Semantic, syntactic, and sequential-based text graphs are constructed to form a text graph tensor, capturing multiple contexts. It encodes heterogeneous information from multi-graphs by performing intra-graph propagation to aggregate node information and inter-graph propagation to harmonize information between graphs.}
    \label{fig9}
\end{figure}
\subsubsection{Multigraph Approaches}
Some approaches leverage multiple graphs to capture additional contextual information from the corpus. Vashishth et al. \cite{c27} proposed capturing additional contextual information, i.e., semantic and syntactic context using multiple graphs. They used two GCNs to independently learn from semantic and syntactic graphs built from the same corpus and demonstrated the efficacy of combining them to perform various tasks. This idea was later built upon and applied to text classification by Liu et al. \cite{c23}. Their proposed approach leveraged a graph tensor that captured semantic, syntactic, and sequential context of textual information using three separate heterogeneous graphs and used a GCN to jointly learn on each of them. This approach simultaneously performed intra-graph propagation to aggregate information from the neighbors of each node and inter-graph propagation to integrate the heterogeneous information across these graphs (Fig. \ref{fig9}).

Wu et al. \cite{c169} proposed a new approach that built corpus-level text graphs instead of defining each document as a node, as that would limit their method to only transductive settings. For document embeddings, their framework merged word embeddings as per document-level structural information in real-time. They built multiple corpus-level graphs to capture different views of structural information, applied graph convolutions to them and fused their results to obtain a better decision boundary. Some works have  leveraged multiple graphs to capture external information not explicitly expressed within the text. Zhou et al. \cite{c171} constructed syntactic and knowledge graphs, combined their adjacency matrices, and applied a GCN to them along with multi-head positional attention to enhance the sentence representation towards a given aspect. Wang et al. \cite{c172} also constructed and fed two graphs based on syntactic dependency and entity relationships into separate GCN modules and fused their outputs to improve Chinese long-text classification performance.

\subsubsection{Inductive Approaches}
While most approaches for text classification are transductive, literature that tries to tackle this problem using inductive processes does exist. Unlike transductive learning techniques in which we have observed all the data beforehand during training, inductive learning relies on only the training data for training the model and then applying the learned model to a dataset it has never seen before.
\begin{figure}[t]
    \centering
    \includegraphics[width=0.9\linewidth]{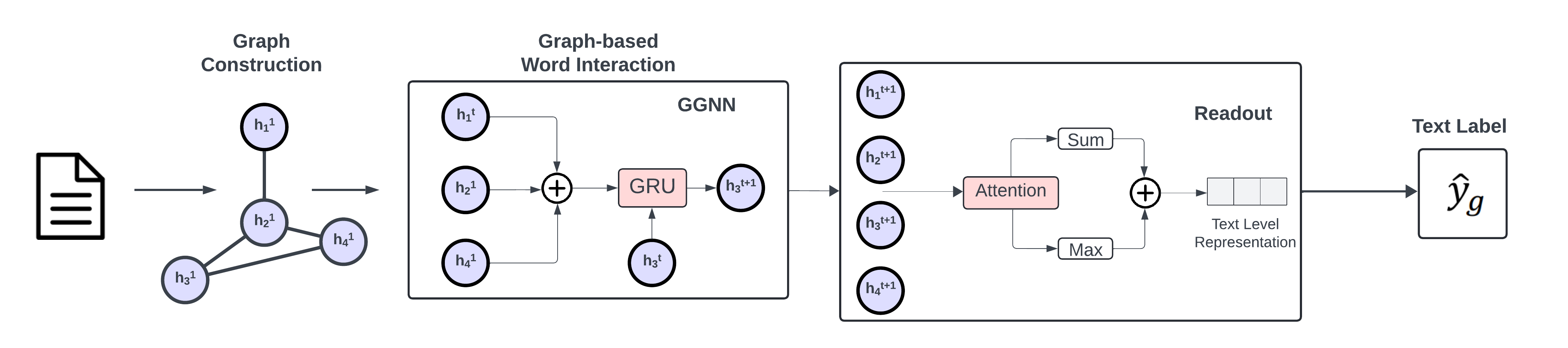}
    \caption{The architecture of TextING, proposed by Zhang et al. \cite{c25}. Separate graphs are constructed for each document, where word feature information is propagated and incorporated contextually during the word interaction phase. Gated Graph Neural Networks are then employed to learn the embeddings of the word nodes. After sufficient updates, word nodes are aggregated to a graph-level representation for the document, based on which the final predictions are made. The readout layer consists of two multilayer perceptrons  applying a soft attention weight, and  a non-linear feature transformation.}
\label{fig10}
\end{figure}

One such approach was proposed by Zhang et al. \cite{c25} to overcome the limitations posed by transductive GCN methods such as TextGCN. 
To learn fine-grained text-level word relations, they first built individual graphs for each document. Information from word nodes was then propagated to their neighbors via Gated GCNs and aggregated into each document embedding, which were in turn used to obtain the final prediction. Their approach achieved good classification performance, however, its most significant gains were underscored under inductive settings (see Fig. \ref{fig10}).
{Wang et al. \cite{c26} also augmented transductive models like TextGCN and SGC with inductive learning. They generated document node representations from one-hot encoded word node vectors weighted by TF-IDF, using only the training set documents. A GCN was then trained with cross-entropy loss on these training document node representations. During testing, unidirectional propagation updated test document nodes by leveraging stored input and hidden layer representations from the training step.}

\subsubsection{Multilabel Classification}
In multilabel text classification, an instance can be assigned multiple number of classes or labels. Cai et al. \cite{c166} featured a hybrid network that uses a pre-trained BERT model to generate context-aware document representations, while a GCN was used to learn contextualized label embeddings. Attention was learned to assign the weight of the label to each word, yielding a label-specific word representation. The context-aware and label-specific word features were then combined and fed into a Bi-LSTM for classification.

Liu and Bin \cite{c167} proposed a GCN and BERT-based framework for multilabel classification on Chinese government hotline event text. A GCN was applied to an abstract meaning representation-based graph  to produce an event topic information vector. It was then fused with an event semantic information vector extracted using BERT to predict the label count. A  memory network stored the event label semantic information and obtained a candidate set, which was then matched with the GCN-BERT fusion vector using an answer selection framework. The top k labels with maximum probability were selected as the output.

Ma et al. \cite{c168} learned label-specific text representations for the documents. However, to achieve this they extracted relevant semantic components for each of the target classes and used a dual graph GCN to model interactions among them based on the statistical label co-occurrence and dynamic reconstruction graph. The resulting component representations were used to predict the document labels.

\subsubsection{Classification with Class Imbalance}
Text classification may sometimes have class imbalance issue. To address this,  a multi-label classification approach for imbalanced clinical text has recently been proposed by He et al. \cite{he2024mcict}. This approach leverages BioBERT, a pre-trained language model specialized for biomedical texts, to obtain fine-grained semantic features. To tackle class imbalance, it incorporates a co-occurrence based embeddings with additional information enhanced GCN, ultimately learning representations. Karahej et al. \cite{c205} proposed Multi-view Minority Class Text Graph Convolutional Network (MMCT-GCN) that addresses minority classes by capturing textual graph representations in addition to sequence-based text representations.

Badiei et al. \cite{c209} combined GCNs and LSTMs to process text, addressing class imbalance with an adversarial loss framework. They used separate weight generators per class to adjust sample weights dynamically during training. Their approach increased weights for misclassified samples and decreased weights for correctly classified ones, enhancing classifier performance over epochs. 
\subsubsection{Extreme Text Classification} Unlike traditional text classification where we might have 10-20 categories, extreme text classification deals with datasets that have hundreds of thousands, or even millions of potential labels \cite{c201}. While there are a vast number of possible labels, there is often a scarcity of training data for each specific label. This means there might be very few examples for some of the more uncommon labels. This type of classification can be useful for tasks like automatically tagging products with highly specific attributes on an e-commerce website, or categorizing scientific research papers within a very granular subject hierarchy. Various approaches to tackle extreme text classification include transfer learning, leveraging pre-trained language models, and developing new methods for handling imbalanced datasets. Transformers are also used for extreme multi-label text classification due to their effective text representation capabilities. Zhao et al. \cite{zhao2024tlc} proposed TLC-XML, a Transformer-based model for XML classification. The model includes three modules: Partition, Matcher, and Ranker. In the Partition module, label correlation graphs are constructed using semantic and co-occurrence information, grouping strongly correlated labels into clusters. The Matcher module employs GCNs for cluster correlation learning, embedding these correlations into the classifier. The Ranker module improves label predictions by integrating raw predictions with information from neighboring labels.

Dahiya et al. \cite{c208} developed the Siamese Extreme Multi-Label GCN model (SiameseXML), leveraging a probabilistic model that supports a modular approach combining Siamese architectures with powerful extreme GCN based classifiers. They also designed a scalable training pipeline capable of handling tasks with up to 100 million labels. Jiang et al. \cite{c206} introduced LightXML, which employed an end-to-end training method along with dynamic sampling of negative labels. 
Xiong et al. \cite{c207} also proposed transformer based two-stage Extreme Multi-label Text Classification (XMTC) model. 




\subsubsection{Multilingual Text Classification} Multilingual text classification involves categorizing written texts in different languages. Multilingual Sentiment Analysis GCN (MSA-GCN) \cite{mercha2024heterogeneous} used a GCN to capture both short-distance and long-distance semantics effectively. This approach employed a unified heterogeneous text graph and uses a moderately deep GCN to acquire predictive representations for all nodes, facilitating transfer learning across languages.
Wu et al. \cite{c170} also used a GCN to capture rich information contained within and across languages for cross-lingual text classification. They added part-of-speech tags to edges as well as direct connections between similar documents and machine-translated versions of the same document in a base text graph. A GCN was then applied to aggregate information from multiple subgraphs separated by different types of edges and learn a language-agnostic representation for the documents.

\subsubsection{Hierarchical Text Classification} Hierarchical text classification is the process of categorizing a text into multiple hierarchical levels of categories. Feng et al. \cite{feng2024adaptive} proposed the Adaptive Micro-knowledge and Macro-Knowledge incorporation for Hierarchical Text Classification (AMKI-HTC) model, which integrated micro-knowledge to capture class-relevant keywords for discriminative representations and enhanced label graph accuracy with macro-knowledge. It incorporated a confidence maximization fusion strategy for adaptive aggregation of multi-view features.

\subsection{Semi-supervised Text Classification}

In a semi-supervised setting, the model is trained using a combination of labeled and unlabeled data. The labeled data guides learning of the relationship between inputs and outputs as it does in supervised learning. However, semi-supervised approaches additionally use unlabeled data to improve the quality of the learned representation so that it is more robust and better generalizes. In practice, this can be achieved by minimizing a loss function that combines the classification loss for the labeled examples with a regularization term that encourages the model to learn similar representations for similar documents.

\subsubsection{Short Text Classification}
TextGCN had previously been extended to short-text classification, namely for product query and product title classification, either by leveraging side information to construct the graph \cite{c162} or by incorporating label dependencies in the output space \cite{c163}. Ye et al. \cite{c164} employed a topic model to obtain global short text topic information to complement the word co-occurrence, and document word relation information for graph construction. A GCN was applied to the short-text graph, and the resulting word and document nodes and pre-trained vector obtained from BERT’s hidden layer were input into a Bi-LSTM classifier.

Yang et al. \cite{c130, c131} proposed inductive learning and multi-label classification of short texts using two steps: first, a flexible Heterogeneous Information Network (HIN) modeled short texts by capturing rich relations among them and augmenting them with additional information, such as topics and entities. Doing so helped alleviate semantic sparsity, combat noise, and make predictions with greater confidence in the downstream classification task. Then, their proposed Heterogeneous Graph Attention (HGAT) model embedded the HIN outputs based on a node and type-level attention mechanism. This dual-level attention mechanism considered the significance of not just neighboring nodes but also the different types of information associated with a particular node. The HGAT employed heterogeneous graph convolution to factor in the difference between various types of information and map them into an implicit shared space using their corresponding transformation matrices.

Zhao et al. \cite{c97} proposed a Multi-head-Pooling-based GCN for more robust short text classification without the need for pre-trained word embeddings and with a lower computational overhead. They introduced three architectures focusing on the first-order nodes of isomorphic graphs, first and second-order nodes of isomorphic graphs, and first-order nodes of heterogeneous graphs, respectively. The key innovation of this approach was the use of a graph pooling method based on self-attention to evaluate and select important nodes from these multiple perspectives without an increase in trainable parameters. 
The authors demonstrated their model’s ability to capture rich semantic information in short texts and effectiveness across multiple benchmark datasets.

Cui et al. \cite{c98} designed a model to overcome the challenges of short text classification due to sparsity and limited labeled data. Instead of generating text samples, they opted for a more convenient self-training method that propagated labeled information to target samples through the graph structure. Their model added keywords to the training set and calculated the confidence of each word. Words with high confidence were identified automatically as pseudo-labeled data, and the confidence of each word was used to compute edge weights in the graph, reducing the impact of ambiguous words on classification performance.

\subsubsection{Multigraph Approaches}
Li et al. \cite{c24} argued against heterogeneous graphs as they increased the number of parameters by an unwarranted amount. Moreover, they claimed that previous graph construction methods relied solely on empirical design, had no theoretical foundations, and introduced additional problems like node redundancy, missing information, and error cascade propagation. Their approach leveraged multiple non-heterogeneous graphs and refined the graph topology to propagate information more effectively. It also incorporated attribute space interpolation based on dense substructure in graphs to predict low-entropy labels with high-quality feature nodes for data augmentation. Overall, they were able to reduce parameter complexity and make the trained model lightweight while still effectively capturing different types of context. 

\subsubsection{Zero-shot Classification}
Liu et al. \cite{c94} proposed a novel method to achieve zero-shot text classification by connecting seen classes to unseen classes using semantic category knowledge from ConceptNet \cite{c203} and constructing a graph of all categories. Unseen classes could then be identified by information propagation through this connection. It was done by transferring category knowledge through convolution on the constructed graph and semi-supervised training using samples of the seen classes. 

\subsubsection{Inductive Approaches}
Ragesh et al. \cite{c95}  addressed the problem of learning efficient and inductive graph convolutional networks for text classification with many examples and features. This work featured a heterogeneous GCN architecture that integrated the best aspects of predictive text embedding (PTE) and TextGCN to derive document embeddings using compatible graphs across multiple layers. It  decomposed TextGCN into simpler models that stored feature embeddings at various layers but had fewer parameters allowing for faster training and better generalization performance when the amount of labeled data was scarce. 

Lin et al. \cite{lin2024heterogeneous} proposed Heterogeneous Directed Graph Attention Networks (HDGAT), which integrates sentence-transformer, global attention mechanism (GAT), and Squeeze-and-Excitation Network (SENet) based channel attention for multilevel semantic embedding and automatic learning of node connections, for text classification

\subsubsection{Document-document Edge Definition}
Wang et al. \cite{c96} improved the performance of text classification by integrating a rich source of graph edge information of the entire text corpus. For the text graph, they used Word2Vec and Doc2Vec embeddings as word and document node features, respectively. They defined document-document edges in addition to word-word and word-document edges. Weights for word-word and document-document edges were inversely proportional to the distance between the feature values of the nodes linking them while word-document edges were represented using TF-IDF values. The generated graph was then trained with their proposed model, which considered the edge features as multi-stream signals, with each stream performing a separate graph convolutional operation. Pooling was used at the output layer to further synthesize the multi-stream features of each node for final classification.
In multi-task text classification, the goal is to classify text data into multiple categories by sharing the knowledge and features learned from different related tasks. Thus, improving the performance of each task by leveraging the similarities and differences among them and reducing the amount of labeled data required for each task. Marreddy et al. \cite{c174} proposed a novel semi-supervised multi-task text classification framework to address the challenges of applying GCN to low-resource languages such as Telugu. It comprised mainly of a graph autoencoder (GAE) and a multi-task GCN. The GAE learned low-dimensional word and sentence graph embeddings from word-sentence graph reconstruction, whereas the multi-task text GCN performed multi-task text classification using these latent sentence graph embeddings. In addition, their approach achieved significant improvements on four text classification tasks, including sentiment analysis, emotion identification, hate speech, and sarcasm detection. 

\subsubsection{Neighborhood-level Contrastive Learning}
Xiao et al. \cite{xiao2024nnc} proposed a simple and efficient Neighbors-to-Neighbors Contrastive GCN (NNC-GCN) for semi-supervised classification. It built consistent multi-views using topologies of original input graphs and used an improved version of Info Noise Contrastive Estimation (InfoNCE) \cite{c202} loss function. InfoNCE was adapted to neighborhood-level contrast learning by weighting and treating the neighborhoods and remaining nodes of the selected anchor as positive and negative sample sets.

\subsection{Self-supervised Text Classification}
In self-supervised conditions, the model is able to learn a meaningful representation of input documents without any explicit supervision through training labels. Instead, the model is trained to predict certain hidden features of the input document based on some other unhidden aspect of the same document. More specifically, the model solves a pretext task from unlabeled data, such as predicting the next sentence in a document or reconstructing a corrupted version of the same document. By doing so, the model can learn to capture important semantic and syntactic features of the documents, which can be useful for downstream tasks such as text classification.

\subsubsection{Multimodal Representation Learning}
The GCNW-FL by Zhu et al. \cite{c101} learned multimodal word representations using GCNs by harnessing their ability to capture the relationships between different language modalities, such as phonetics and syntax. To train their model, they used a greedy strategy to update the modality-relation matrix in the GCN and effectively learn multimodal word representations by predicting the context of words from their phonetic and syntactic information. They evaluated it on downstream text classification task, and demonstrated its efficacy at capturing rich semantics through the learned word representations. 

\subsubsection{Contrastive Learning with Augmentation Strategies}
Yang et al. \cite{c99} obtained a robust node representation through contrastive learning using noise and centrality-based augmentations. This enabled them to preserve essential connections between nodes while also reducing noise at the same time. They used nodes with the same label as multiple positive samples and assigned them to the anchor node while applying consistency training on unlabeled nodes to constrain model predictions. They used random node sampling for more efficient resource utilization while computing the contrastive loss.

\subsubsection{Pre-trained Language Model Integration}
Similar to previously discussed inductive methods \cite{c25, c26}, Wu et al. \cite{c100} addressed the limitation posed by the transductive nature of most GCN-based models and the challenge of deploying a GCN-based model in an online system, where new data is added continually, and the model needs to be updated to account for this change. They proposed a new `all-token-any-document (ATAD)' paradigm that uses the vocabulary of a pre-trained language model such as BERT \cite{c109}, RoBERTa \cite{c204}, or XLNet \cite{c200} to dynamically update the connections between documents and tokens in the graph, allowing the model to predict for previously unseen documents. They introduced a method for online updating without the need for labels. This approach fine-tuned an occurrence memory module and efficiently updated the network parameters using a self-supervised contrastive learning objective. 

\subsection{Weakly Supervised Approaches}
Weak supervision is a form of machine learning where the model is trained using limited, or imprecise labels. Instead of having access to a fully labeled dataset, weak supervision leverages various forms of indirect supervision to construct approximate labels. These forms can include domain knowledge, heuristics, and other semi-automated methods. Weakly supervision is particularly useful in scenarios where obtaining fully labeled data is challenging, such as in large-scale datasets or specialized domains like  natural language processing.

\subsubsection{Multiple Instance Learning}
Multiple Instance Learning (MIL) is a weakly supervised learning framework where the model receives a set of labeled bags, each containing multiple instances. The label is provided at the bag level, not the instance level, which means that the model must learn to predict the bag label based on the instances within it. MIL is particularly effective in scenarios where instance-level labels are not available but bag-level labels are. 
The MIL framework naturally fits various problem settings and as a result, it has been applied to various domains such as computer vision \cite{r18, r19, r20}, natural language processing \cite{r1, r21, r22, r23}, anomaly detection \cite{r29, r30}, remote sensing \cite{r27, r28}, and medical image analysis \cite{r24, r25, r26}. In text classification, MIL can be applied by considering documents as bags and sentences or paragraphs as instances. The goal is to classify the entire document (bag) based on the information contained within its sentences or paragraphs (instances). Traditional MIL approaches often treat instances as independent and identically distributed (i.i.d.), which overlooks the structural relationships between them. Thus, the combination of MIL with GCNs can be potentially advantageous in applications where preserving complex relationships between instances within bags is essential, such as text categorization \cite{c211}, medical imaging \cite{r35}, and speech classification \cite{r36}.

Tu et al. \cite{c211} integrated MIL with GCNs by treating each bag as a graph, with instances as nodes connected by edges representing their relationships. A GCN was applied to this graph to learn node embeddings that captured the structural information of the instances. These embeddings were then aggregated to form a representation of the entire bag, which was used by a classifier to predict the bag-level label. By capturing the structural information within the bag, this method provided a more nuanced representation, improving classification performance.

\subsection{Discussion}
{
The landscape of GCN approaches for text classification has evolved significantly, marked by a transition from traditional supervised methods to more sophisticated semi-supervised and self-supervised techniques. Initially, supervised approaches dominated the field, focusing on leveraging labeled data to train models. These early models, such as TextGCN \cite{c17}, TL-GNN \cite{c18} and SGC \cite{c19}, aimed to optimize computational efficiency and reduce complexity while enhancing classification performance.
}

{As the field progressed, researchers explored multigraph approaches like TensorGCN \cite{c23}, which utilized multiple graphs to capture richer contextual information, such as semantic and syntactic structures. This shift allowed models to better understand nuanced text relationships, leading to improved classification accuracy. Inductive approaches such as TextING \cite{c25} and InducT-GCN \cite{c26} also emerged to address the limitations of transductive methods, enhancing the models' ability to generalize to unseen data without retraining. The development of multilabel classification techniques, including HBLA \cite{c166} and GCN-BERT \cite{c167}, marked another significant advancement. These models handled the complexity of assigning multiple labels to a single instance, addressing challenges like label correlation and class imbalance. Recent innovations introduced specialized subcategories, like multilingual models (e.g., MSA-GCN \cite{mercha2024heterogeneous}) and hierarchical models (e.g., AMKI-HTC \cite{feng2024adaptive}), expanding the scope of GCN applications to more complex and varied classification tasks.
}

{In subsequent years, there was a noticeable shift towards semi-supervised methods, motivated by the need to leverage large amounts of unlabeled data, which is often more readily available than labeled datasets. Notably, HeteGCN \cite{c95} demonstrated consistent high performance across various datasets, solidifying the importance of semi-supervised approaches in achieving robust and scalable text classification. More recently, the research focus has shifted towards self-supervised learning, reflecting a broader trend in machine learning towards minimizing the dependency on labeled data. Advanced architectures like Cont-GCN-BERT \cite{c100} represent the cutting edge research, utilizing self-generated supervisory signals to train models. These methods have shown remarkable performance by integrating self-supervised learning with powerful pre-trained language models.
}

{Overall, there is a consistent trend of performance improvement across benchmarks. Early models focused on addressing basic computational and memory constraints, while later models incorporated more complex architectural innovations to capture richer contextual information and handle diverse classification challenges effectively. The semi-supervised and self-supervised approaches reflect a growing emphasis on scalability and adaptability, allowing GCNs to be applied to larger and more varied datasets without relying on labels.
}

\section{Performance Comparison}
{
To evaluate and compare different GCN-based approaches, we adopted a rigorous methodology:
\begin{itemize}
    \item \textbf{Selection Criteria}: Approaches were selected based on their relevance, innovation, impact, and citations. Both foundational and recent high-impact works are included.
    \item \textbf{Datasets}: We utilized widely recognized benchmark datasets to ensure consistent and meaningful comparisons across methods. Details of these datasets are provided in Section \ref{datasets}.
    \item \textbf{Metrics}: While accuracy scores are the most abundantly reported metric across all approaches, macro-averaged F1 scores have also been considered wherever possible to highlight notable trends across different categories. We primarily reported test accuracy and sometimes test F1 scores in our analysis, but a more holistic discussion on metrics is provided in Section \ref{metrics}.
    \item \textbf{Experimental Setup}: We reviewed the experimental setups reported in the literature to ensure comparisons are fair, considering factors like training data, model configurations, and evaluation protocols. Some methods used different train-test splits, which we have reported accordingly.
    \item \textbf{Comparative Analysis}: Comparisons have been made by evaluating approaches within each category, namely supervised, semi-supervised, and self-supervised.
\end{itemize}
}

\subsection{Datasets}
\label{datasets}
{In this section, we provide specifics on some of the most widely used datasets in relevant literature that have been used to benchmark the text classification performance of GCN methods. As results on these datasets have been reported across numerous studies, we can use them to provide a more meaningful comparison of approaches to highlight their relative strengths and limitations. Their statistics in the standard configuration \cite{c17} are also summarized in Table \ref{tab3}.}

\begin{itemize}

\item \textbf{20 NG:} The 20 NG dataset contains 18,846 newsgroup documents evenly categorized into 20 different categories, covering a broad spectrum of topics such as sports, politics, technology, and religion, among others. In total, 11,314 documents are in the training set and 7,532 documents are in the test set. In addition to text classification, this dataset has also been used for text clustering and out-of-distribution detection.
\item \textbf{Reuters:} This dataset is a collection of documents that appeared on Reuters newswire in 1987 and has been widely used for text classification. R8 and R52 are popular subsets of the Reuters dataset. The former has 8 news categories split into 5,485 training and 2,189 test documents while the latter contains 52 categories, split into 6,532 training and 2,568 test documents.
\item \textbf{Movie Review (MR):} The MR dataset is comprised of movie reviews and used primarily for sentiment analysis, i.e., whether a review is negative or positive. It contains 5,331 positive and 5,331 negative reviews.
\item \textbf{Ohsumed:} Ohsumed collects medical abstracts tagged by one or multiple classes from 23 cardiovascular disease categories from the MEDLINE database. Since most literature focuses on single-class classification, out of these only 7,400 single-category documents are retained, out of which 3,357 make up the training set and 4,043 documents are in the test set.
\item \textbf{CoLA:} The CoLA (Corpus of Linguistic Acceptability) dataset is a collection of English sentences labeled for grammatical acceptability. The labels are binary (i.e., a sentence is either linguistically acceptable or it is not). This dataset was introduced by Warstadt et al. \cite{c82} and consists of 10,000 sentences from various sources, including linguistic literature, standardized tests, and online forums. The publically available version has 9594 sentences in training and development sets and 1063 sentences in the test set.
\item \textbf{SST-2:} First introduced by Socher et al. \cite{c83}, this variant of the Stanford Sentiment Treebank (SST) dataset is a collection of movie reviews labeled with binary sentiment classification (positive or negative). The reviews are parsed into constituency trees and labeled with sentiment annotations for each sub-phrase, allowing for fine-grained analysis of the sentiment of the text. The dataset consists of over 11,000 sentences from movie reviews and is widely used as a benchmark dataset for evaluating the performance of models on text classification tasks.
\end{itemize}

\begin{table*}[t]
\centering
\resizebox{0.7\textwidth}{!}{%
\begin{tabular}{|l|l|l|l|l|l|l|l|l|}
\hline
\textbf{Dataset} & \textbf{Year} & \textbf{Docs}   & \textbf{Train} & \textbf{Test}  & \textbf{Words}  & \textbf{Nodes}  & \textbf{Classes} & \textbf{Avg. Length} \\ \hline
20 NG & 1995 & 18,846 & 11,314 & 7,532 & 42,757 & 61,603 & 20 & 221.3          \\ \hline
R8 & 2004 & 7,674  & 5,485 & 2,189 & 7,688  & 15,362 & 8 & 65.7           \\ \hline
R52 & 2004 & 9,100  & 6,532 & 2,568 & 8,892 & 17,992 & 52    & 69.8           \\ \hline
Ohsumed & 1994 & 7,400  & 3,357    & 4,043 & 14,157 & 21,557 & 23      & 135.8          \\ \hline
MR & 2005  & 10,662 & 7,108    & 3,554 & 18,764 & 29,426 & 2       & 20.4           \\ \hline
CoLA \cite{c82} & 2019  & 9,594  & 8,551    & 1,043 & -      & -      & 2       & 7.7            \\ \hline
SST-2 \cite{c83} & 2013   & 9,613  & 7,792    & 1,821 & -      & -      & 2       & 19.3           \\ \hline        
\end{tabular}%
}
\caption{\small Summary statistics of benchmark datasets.}
\label{tab3}
\end{table*}

These datasets have been used as benchmarks for a plethora of applications in addition to text classification. Some of these have been summarized in Table \ref{tab4}.

\begin{table*}[t]
\centering
\resizebox{0.7\textwidth}{!}{%
\begin{tabular}{|l|l|l|}
\hline
\textbf{Dataset}                       & \textbf{Task}                      & \textbf{Notable architectures}                                    \\ \hline
\multirow{2}{*}{20 NG}        & Text Clustering, Topic Modelling & G-BAT \cite{c84}                                                    \\ \cline{2-3} 
                              & Out-of-distribution Detection    & 2-Layered GRU \cite{c85}                                           \\ \hline
Reuters                       & Multi-label Text Classification  & HiddeN \cite{c86}                                                   \\ \hline
\multirow{2}{*}{Movie Review} & Sentiment Analysis               & VLAWE \cite{c87}, EFL \cite{c88}                                               \\ \cline{2-3} 
                              & Few-shot Learning                & DART \cite{c89}                                                     \\ \hline
Ohsumed                       & Information Retrieval            & BERT+CONCEPT FILTER \cite{c90}                                      \\ \hline
CoLA                          & Linguistic Acceptability         & \begin{tabular}[c]{@{}l@{}}DeBERTa \cite{c91}, EFL\end{tabular} \\ \hline
\multirow{2}{*}{SST-2}        & Sentiment Analysis               & T5-11B \cite{c92}, MT-DNN-SMART \cite{c93}                                     \\ \cline{2-3} 
                              & Few-shot Learning                & DART                                                     \\ \hline
\end{tabular}
}
\caption{\small Non-text classification tasks performed on selected benchmark datasets with notable architectures.}
\label{tab4}
\end{table*}

\subsection{Metrics}
\label{metrics}

This section serves as a primer for various metrics that have been used for text classification in literature. In the definitions that follow $TP$, $FP$, $TN$, $FN$ have been used to denote true positive, false positive, true negative, and false negative,
respectively: Accuracy = (TP+TN)/(TP+FP+FN+TN), 
Precision = {TP}/(TP+FP), Recall = {TP}/{(TP+FN)},
F1-score = {(2.Recall.Precision)}/{(Recall + Precision)}.
In literature, primarily the test accuracy has been used to evaluate the performance of various models. However,  F1-score, Precision, and Recall are also used to evaluate an approach depending on the exact nature of the application and distribution of a dataset.

In general, accuracy is a good measure when we have a balanced class distribution, whereas the F1-score is usually best for imbalanced classes. Precision is preferable when the goal is to optimize for True Positives while one should choose Recall if the occurrence of false positives is more desirable than the occurrence of false negatives. The F1-score represents a balance of sorts between Precision and Recall as it is derived from their harmonic mean. For multi-class classification problems, most text classification methods  compute precision, recall, or F1-score for each class and compute its  macro average. Such averaging considers each class equally important and balances out the effect of majority classes.  

\subsection{Analysis of Results}
This section compares the performance of various methods discussed in this review over a range of benchmark datasets. Primarily, comparisons have been made by evaluating approaches within each category, namely supervised, semi-supervised, and self-supervised. However, cross-category comparisons have also been presented for a more thorough understanding of the relative strengths and weaknesses of these approaches. While accuracy scores are the most abundantly reported metric across all of these approaches, macro-averaged F1-scores have also been considered wherever possible to highlight any notable trends across different categories. Metrics for all GCN methods have been categorically presented under different conditions in Tables \ref{tab5} to \ref{tab7}. For each of these tables, if a set of results has been obtained under a different train/test split than in \cite{c17}, its details can be found in the table notes. If for a given split, results have been obtained from papers other than the original paper of that approach, those have also been cited next to the method name.

\begin{table*}[]
\centering
\resizebox{0.8\textwidth}{!}{
\begin{threeparttable}[b]
\caption{\small Test accuracies and macro-averaged F1 scores for various supervised text classification approaches on seven benchmarks.}
\centering
\begin{tabular}{|p{2.5cm}|l|ll|ll|ll|ll|ll|ll|ll|}
\hline
\textbf{Method}        & \textbf{Year} & \multicolumn{2}{l|}{\textbf{20 NG}}              & \multicolumn{2}{l|}{\textbf{R8}}                 & \multicolumn{2}{l|}{\textbf{R52}}                & \multicolumn{2}{l|}{\textbf{Ohsumed}}            & \multicolumn{2}{l|}{\textbf{MR}}                 & \multicolumn{2}{l|}{\textbf{CoLA}}               & \multicolumn{2}{l|}{\textbf{SST-2}}              \\ \hline
\textbf{}              & \textbf{}     & \multicolumn{1}{l|}{\textbf{Acc.}} & \textbf{F1} & \multicolumn{1}{l|}{\textbf{Acc.}} & \textbf{F1} & \multicolumn{1}{l|}{\textbf{Acc.}} & \textbf{F1} & \multicolumn{1}{l|}{\textbf{Acc.}} & \textbf{F1} & \multicolumn{1}{l|}{\textbf{Acc.}} & \textbf{F1} & \multicolumn{1}{l|}{\textbf{Acc.}} & \textbf{F1} & \multicolumn{1}{l|}{\textbf{Acc.}} & \textbf{F1} \\ \hline
TextGCN \cite{c17}     & 2019          & \multicolumn{1}{l|}{86.3}          & 85.6        & \multicolumn{1}{l|}{97.1}          & 92.4        & \multicolumn{1}{l|}{93.6}          & 65.2        & \multicolumn{1}{l|}{68.4}          & 59.1        & \multicolumn{1}{l|}{76.7}          & 76.8        & \multicolumn{1}{l|}{52.3}          & 52.3        & \multicolumn{1}{l|}{82.4}          & 80.5        \\ \hline
TextGCN\tnote{1} \hspace{1sp} \cite{c17}               & 2019          & \multicolumn{1}{l|}{77.6}          &             & \multicolumn{1}{l|}{85.6}          &             & \multicolumn{1}{l|}{81.4}          &             & \multicolumn{1}{l|}{50.9}          &             & \multicolumn{1}{l|}{60.5}          &             & \multicolumn{1}{l|}{}              &             & \multicolumn{1}{l|}{}              &             \\ \hline
TextGCN\tnote{3} \hspace{1sp} \cite{c17}            & 2019          & \multicolumn{1}{l|}{80.9}          & 80.5        & \multicolumn{1}{l|}{94.0}          & 78.3        & \multicolumn{1}{l|}{89.4}          & 47.3        & \multicolumn{1}{l|}{56.3}          & 36.7        & \multicolumn{1}{l|}{74.6}          & 74.5        & \multicolumn{1}{l|}{}              &             & \multicolumn{1}{l|}{}              &             \\ \hline
TextGCN\tnote{4} \hspace{1sp} \cite{c17}             & 2019          & \multicolumn{1}{l|}{11.9}          &             & \multicolumn{1}{l|}{86.3}          &             & \multicolumn{1}{l|}{48.5}          &             & \multicolumn{1}{l|}{16.1}          &             & \multicolumn{1}{l|}{62.2}          &             & \multicolumn{1}{l|}{}              &             & \multicolumn{1}{l|}{}              &             \\ \hline
TextGCN\tnote{5} \hspace{1sp} \cite{c17}          & 2019          & \multicolumn{1}{l|}{}              &             & \multicolumn{1}{l|}{}              &             & \multicolumn{1}{l|}{}              &             & \multicolumn{1}{l|}{40.2}          & 23.2        & \multicolumn{1}{l|}{70.3}          & 70.2        & \multicolumn{1}{l|}{}              &             & \multicolumn{1}{l|}{}              &             \\ \hline
TextGCN\tnote{6} \hspace{1sp} \cite{c17}          & 2019          & \multicolumn{1}{l|}{}              &             & \multicolumn{1}{l|}{}              &             & \multicolumn{1}{l|}{}              &             & \multicolumn{1}{l|}{41.6}          & 27.4        & \multicolumn{1}{l|}{59.1}          & 59.0        & \multicolumn{1}{l|}{}              &             & \multicolumn{1}{l|}{}              &             \\ \hline
TextGCN\tnote{7} \hspace{1sp} \cite{c17}          & 2019          & \multicolumn{1}{l|}{}              &             & \multicolumn{1}{l|}{91.2}              &             & \multicolumn{1}{l|}{78.9}              &             & \multicolumn{1}{l|}{22.3}          &         & \multicolumn{1}{l|}{53.4}          &         & \multicolumn{1}{l|}{}              &             & \multicolumn{1}{l|}{}              &             \\ \hline
SGC \cite{c19}         & 2019          & \multicolumn{1}{l|}{88.5}          &             & \multicolumn{1}{l|}{97.2}          &             & \multicolumn{1}{l|}{94.0}          &             & \multicolumn{1}{l|}{68.5}          &             & \multicolumn{1}{l|}{75.9}          &             & \multicolumn{1}{l|}{}              &             & \multicolumn{1}{l|}{}              &             \\ \hline
SGC\tnote{7} \hspace{1sp} \cite{c19}          & 2019          & \multicolumn{1}{l|}{}              &             & \multicolumn{1}{l|}{89.6}              &             & \multicolumn{1}{l|}{77.3}              &             & \multicolumn{1}{l|}{24.7}          &         & \multicolumn{1}{l|}{60.2}          &         & \multicolumn{1}{l|}{}              &             & \multicolumn{1}{l|}{}              &             \\ \hline
TL-GNN \cite{c18}      & 2019          & \multicolumn{1}{l|}{85.9}          &             & \multicolumn{1}{l|}{97.8}          & 95.9        & \multicolumn{1}{l|}{94.6}          & 92.5        & \multicolumn{1}{l|}{69.4}          & 54.0        & \multicolumn{1}{l|}{76.4}          & 76.1        & \multicolumn{1}{l|}{}              &             & \multicolumn{1}{l|}{}              &             \\ \hline
VGCN-BERT \cite{c21}   & 2020          & \multicolumn{1}{l|}{55.8}          &             & \multicolumn{1}{l|}{98.0}          & 95.4        & \multicolumn{1}{l|}{95.9}          &             & \multicolumn{1}{l|}{70.2}          &             & \multicolumn{1}{l|}{86.4}          & 86.4        & \multicolumn{1}{l|}{83.7}          & 80.5        & \multicolumn{1}{l|}{91.9}          & 91.9        \\ \hline
TensorGCN \cite{c23}   & 2020          & \multicolumn{1}{l|}{87.7}          &             & \multicolumn{1}{l|}{98.0}          &             & \multicolumn{1}{l|}{95.1}          &             & \multicolumn{1}{l|}{70.1}          &             & \multicolumn{1}{l|}{77.9}          &             & \multicolumn{1}{l|}{}              &             & \multicolumn{1}{l|}{}              &             \\ \hline
TensorGCN\tnote{1} \hspace{1sp} \cite{c23}            & 2020          & \multicolumn{1}{l|}{78.6}          &             & \multicolumn{1}{l|}{86.2}          &             & \multicolumn{1}{l|}{82.3}          &             & \multicolumn{1}{l|}{52.2}          &             & \multicolumn{1}{l|}{61.3}          &             & \multicolumn{1}{l|}{}              &             & \multicolumn{1}{l|}{}              &             \\ \hline
TextING \cite{c25}     & 2020          & \multicolumn{1}{l|}{82.5}          &             & \multicolumn{1}{l|}{98.1}          &             & \multicolumn{1}{l|}{95.7}          &             & \multicolumn{1}{l|}{70.8}          &             & \multicolumn{1}{l|}{80.2}          &             & \multicolumn{1}{l|}{}              &             & \multicolumn{1}{l|}{}              &             \\ \hline
TextING\tnote{5} \hspace{1sp} \cite{c25}           & 2020          & \multicolumn{1}{l|}{}              &             & \multicolumn{1}{l|}{}              &             & \multicolumn{1}{l|}{}              &             & \multicolumn{1}{l|}{41.8}          & 24.9        & \multicolumn{1}{l|}{69.9}          & 69.7        & \multicolumn{1}{l|}{}              &             & \multicolumn{1}{l|}{}              &             \\ \hline
TextING\tnote{7} \hspace{1sp} \cite{c26}         & 2020          & \multicolumn{1}{l|}{}              &             & \multicolumn{1}{l|}{86.5}              &             & \multicolumn{1}{l|}{74.7}              &             & \multicolumn{1}{l|}{30.3}          &         & \multicolumn{1}{l|}{61.2}          &         & \multicolumn{1}{l|}{}              &             & \multicolumn{1}{l|}{}              &             \\ \hline   
SSGC \cite{c20}        & 2021          & \multicolumn{1}{l|}{88.6}          &             & \multicolumn{1}{l|}{97.4}          &             & \multicolumn{1}{l|}{94.5}          &             & \multicolumn{1}{l|}{68.5}          &             & \multicolumn{1}{l|}{76.7}          &             & \multicolumn{1}{l|}{}              &             & \multicolumn{1}{l|}{}              &             \\ \hline
BERT-GCN \cite{c22}    & 2021          & \multicolumn{1}{l|}{89.3}          &             & \multicolumn{1}{l|}{98.1}          &             & \multicolumn{1}{l|}{96.6}          &             & \multicolumn{1}{l|}{72.8}          &             & \multicolumn{1}{l|}{86.0}          &             & \multicolumn{1}{l|}{}              &             & \multicolumn{1}{l|}{}              &             \\ \hline
RoBERTa-GCN \cite{c22} & 2021          & \multicolumn{1}{l|}{89.5}          &             & \multicolumn{1}{l|}{98.2}          &             & \multicolumn{1}{l|}{96.1}          &             & \multicolumn{1}{l|}{72.8}          &             & \multicolumn{1}{l|}{89.7}          &             & \multicolumn{1}{l|}{}              &             & \multicolumn{1}{l|}{}              &             \\ \hline
NMGC-2 \cite{c173} & 2021          & \multicolumn{1}{l|}{86.6}          &             & \multicolumn{1}{l|}{97.3}          &             & \multicolumn{1}{l|}{94.4}          &             & \multicolumn{1}{l|}{69.2}          &             & \multicolumn{1}{l|}{76.2}          &             & \multicolumn{1}{l|}{}              &             & \multicolumn{1}{l|}{}              &             \\ \hline
TGCN-Bi-LSTM \cite{c182} & 2021          & \multicolumn{1}{l|}{93.0}          & {92.7}            & \multicolumn{1}{l|}{97.6}          &  {94.0}           & \multicolumn{1}{l|}{94.7}          & {72.9}            & \multicolumn{1}{l|}{72.2}          & {68.4}            & \multicolumn{1}{l|}{}          &             & \multicolumn{1}{l|}{}              &             & \multicolumn{1}{l|}{}              &             \\ \hline
TGCN-C-LSTM \cite{c182} & 2021          & \multicolumn{1}{l|}{93.2}          & {93.0}            & \multicolumn{1}{l|}{97.6}          &  {93.6}           & \multicolumn{1}{l|}{94.4}          & {71.3}            & \multicolumn{1}{l|}{72.6}          & {68.4}            & \multicolumn{1}{l|}{}          &             & \multicolumn{1}{l|}{}              &             & \multicolumn{1}{l|}{}              &             \\ \hline
TGCN-ServeNet \cite{c182} & 2021          & \multicolumn{1}{l|}{92.9}          & {92.7}            & \multicolumn{1}{l|}{97.9}          &  {94.6}           & \multicolumn{1}{l|}{94.9}          & {74.2}            & \multicolumn{1}{l|}{72.1}          & {68.4}            & \multicolumn{1}{l|}{}          &             & \multicolumn{1}{l|}{}              &             & \multicolumn{1}{l|}{}              &             \\ \hline
MGCN \cite{c175} & 2021          & \multicolumn{1}{l|}{}          &             & \multicolumn{1}{l|}{}          &             & \multicolumn{1}{l|}{}          &             & \multicolumn{1}{l|}{}          &             & \multicolumn{1}{l|}{}          & 87.4            & \multicolumn{1}{l|}{}              & 80.3             & \multicolumn{1}{l|}{} & 92.3                         \\ \hline
GFN \cite{c169} & 2022          & \multicolumn{1}{l|}{87.0}          & {86.3}            & \multicolumn{1}{l|}{98.2}          &  {95.5}           & \multicolumn{1}{l|}{95.3}          & {74.6}            & \multicolumn{1}{l|}{70.2}          & {60.3}            & \multicolumn{1}{l|}{78.0}          & {77.8}            & \multicolumn{1}{l|}{}              &             & \multicolumn{1}{l|}{}              &             \\ \hline
IMGCN \cite{c153} & 2022          & \multicolumn{1}{l|}{}          &             & \multicolumn{1}{l|}{98.3}          &   96.5          & \multicolumn{1}{l|}{}          &             & \multicolumn{1}{l|}{}          &             & \multicolumn{1}{l|}{87.8}          &             & \multicolumn{1}{l|}{84.4}              &      80.9        & \multicolumn{1}{l|}{92.5} &                          \\ \hline
GCN-CNN \cite{c187} & 2022          & \multicolumn{1}{l|}{}          &             & \multicolumn{1}{l|}{98.5}          &             & \multicolumn{1}{l|}{96.4}          &             & \multicolumn{1}{l|}{71.9}          &             & \multicolumn{1}{l|}{87.6}          &             & \multicolumn{1}{l|}{}              &             & \multicolumn{1}{l|}{}              &             \\ \hline
BiGRU+GCN \cite{c184} & 2022          & \multicolumn{1}{l|}{86.8}          & {85.5}            & \multicolumn{1}{l|}{97.1}          &  {93.4}           & \multicolumn{1}{l|}{93.9}          & {70.7}            & \multicolumn{1}{l|}{68.4}          & {62.2}            & \multicolumn{1}{l|}{77.6}          & {77.5}            & \multicolumn{1}{l|}{}              &             & \multicolumn{1}{l|}{}              &             \\ \hline
InducT-GCN \cite{c26}  & 2022          & \multicolumn{1}{l|}{}              &             & \multicolumn{1}{l|}{96.5}          & 95.4        & \multicolumn{1}{l|}{93.2}          & 92.8        & \multicolumn{1}{l|}{67.8}          & 66.9        & \multicolumn{1}{l|}{75.4}          & 75.9        & \multicolumn{1}{l|}{}              &             & \multicolumn{1}{l|}{}              &             \\ \hline
InducT-SGC\tnote{7} \hspace{1sp} \cite{c26}          & 2022          & \multicolumn{1}{l|}{}              &             & \multicolumn{1}{l|}{90.5}              &             & \multicolumn{1}{l|}{80.5}              &             & \multicolumn{1}{l|}{31.1}          &         & \multicolumn{1}{l|}{60.2}          &         & \multicolumn{1}{l|}{}              &             & \multicolumn{1}{l|}{}              &             \\ \hline
InducT-GCN\tnote{7} \hspace{1sp} \cite{c26}          & 2022          & \multicolumn{1}{l|}{}              &             & \multicolumn{1}{l|}{91.6}              &             & \multicolumn{1}{l|}{81.4}              &             & \multicolumn{1}{l|}{35.6}          &         & \multicolumn{1}{l|}{60.4}          &         & \multicolumn{1}{l|}{}              &             & \multicolumn{1}{l|}{}              &             \\ \hline
HINT-G \cite{c179} & 2023          & \multicolumn{1}{l|}{87.7}          &             & \multicolumn{1}{l|}{98.2}          &            & \multicolumn{1}{l|}{95.0}          &             & \multicolumn{1}{l|}{72.7}          &             & \multicolumn{1}{l|}{78.2}          &             & \multicolumn{1}{l|}{}              &             & \multicolumn{1}{l|}{}              &             \\ \hline
GTG \cite{c188} & 2023          & \multicolumn{1}{l|}{87.0}          & {85.7}            & \multicolumn{1}{l|}{97.2}          &  {93.7}           & \multicolumn{1}{l|}{94.5}          & {71.2}            & \multicolumn{1}{l|}{69.7}          & {62.8}            & \multicolumn{1}{l|}{77.2}          & {77.0}            & \multicolumn{1}{l|}{}              &             & \multicolumn{1}{l|}{}              &             \\ \hline
LDGCN \cite{c129}      & 2023          & \multicolumn{1}{l|}{87.8}          & 87.8        & \multicolumn{1}{l|}{98.3}          & 97.3        & \multicolumn{1}{l|}{95.7}          & 92.7        & \multicolumn{1}{l|}{70.9}          & 59.1        & \multicolumn{1}{l|}{78.3}          & 78.2        & \multicolumn{1}{l|}{}              &             & \multicolumn{1}{l|}{}              &             \\ \hline

\end{tabular}
\begin{tablenotes}
    \item [1] 20 labeled data per class as reported in \cite{c24}
    \item [3] 20\% stratified sample of training documents as reported in \cite{c95}
    \item [4] 1-99 train/test split as reported in \cite{c96}
    \item [5] 10-90 train/test split as reported in \cite{c98}
    \item [6] 40 labeled data per class as reported in \cite{c131, c99}
    \item [7] 5-95 train/test split as reported in \cite{c26}
 
\end{tablenotes}
\label{tab5}
\end{threeparttable}
}
\end{table*}

\begin{table*}[]
\centering
\resizebox{0.8\textwidth}{!}{
\begin{threeparttable}[b]
\caption{\small Test accuracies and macro-averaged F1 scores for various semi-supervised approaches under different conditions.}
\begin{tabular}{|l|l|ll|ll|ll|ll|ll|}
\hline
\textbf{Method}                & \textbf{Year} & \multicolumn{2}{l|}{\textbf{20 NG}}              & \multicolumn{2}{l|}{\textbf{R8}}                 & \multicolumn{2}{l|}{\textbf{R52}}                & \multicolumn{2}{l|}{\textbf{Ohsumed}}            & \multicolumn{2}{l|}{\textbf{MR}}                 \\ \hline
\textbf{}                      & \textbf{}     & \multicolumn{1}{l|}{\textbf{Acc.}} & \textbf{F1} & \multicolumn{1}{l|}{\textbf{Acc.}} & \textbf{F1} & \multicolumn{1}{l|}{\textbf{Acc.}} & \textbf{F1} & \multicolumn{1}{l|}{\textbf{Acc.}} & \textbf{F1} & \multicolumn{1}{l|}{\textbf{Acc.}} & \textbf{F1} \\ \hline
STGCN \cite{c164}              & 2020          & \multicolumn{1}{l|}{}          &             & \multicolumn{1}{l|}{97.2}          &             & \multicolumn{1}{l|}{}          &             & \multicolumn{1}{l|}{}          &             & \multicolumn{1}{l|}{78.2}          &             \\ \hline
STGCN+BiLSTM \cite{c164}              & 2020          & \multicolumn{1}{l|}{86.6}          &   85.4          & \multicolumn{1}{l|}{97.4}          &    94.3         & \multicolumn{1}{l|}{94.2}          &   71.0          & \multicolumn{1}{l|}{69.2}          & 62.3            & \multicolumn{1}{l|}{78.5}          &    78.2         \\ \hline
STGCN+BERT+BiLSTM \cite{c164}              & 2020          & \multicolumn{1}{l|}{}          &             & \multicolumn{1}{l|}{98.5}          &             & \multicolumn{1}{l|}{}          &             & \multicolumn{1}{l|}{}          &             & \multicolumn{1}{l|}{82.5}          &             \\ \hline
TextGTL\tnote{1} \hspace{1sp} \cite{c24}            & 2021          & \multicolumn{1}{l|}{80.1}          &             & \multicolumn{1}{l|}{87.1}          &             & \multicolumn{1}{l|}{83.2}          &             & \multicolumn{1}{l|}{54.1}          &             & \multicolumn{1}{l|}{62.4}          &             \\ \hline
ZS-TC\tnote{2} \hspace{1sp} \cite{c94}             & 2021          & \multicolumn{1}{l|}{69.0}          &             & \multicolumn{1}{l|}{}              &             & \multicolumn{1}{l|}{}              &             & \multicolumn{1}{l|}{}              &             & \multicolumn{1}{l|}{}              &             \\ \hline
HeteGCN (F-X)\tnote{3} \hspace{1sp} \cite{c95}    & 2021          & \multicolumn{1}{l|}{84.6}          & 84.0        & \multicolumn{1}{l|}{97.2}          & 92.3        & \multicolumn{1}{l|}{93.9}          & 66.5        & \multicolumn{1}{l|}{63.8}          & 50.2        & \multicolumn{1}{l|}{75.6}          & 75.6        \\ \hline
HeteGCN (X-TX-X)\tnote{3} \hspace{1sp} \cite{c95} & 2021          & \multicolumn{1}{l|}{84.1}          & 83.4        & \multicolumn{1}{l|}{97.3}          & 92.9        & \multicolumn{1}{l|}{93.2}          & 61.3        & \multicolumn{1}{l|}{65.3}          & 54.0        & \multicolumn{1}{l|}{75.5}          & 75.5        \\ \hline
HeteGCN (TX-X)\tnote{3} \hspace{1sp} \cite{c95}   & 2021          & \multicolumn{1}{l|}{84.8}          & 84.3        & \multicolumn{1}{l|}{97.1}          & 92.0        & \multicolumn{1}{l|}{93.7}          & 66.0        & \multicolumn{1}{l|}{65.8}          & 57.1        & \multicolumn{1}{l|}{76.1}          & 76.1        \\ \hline
HeteGCN (F-X) \cite{c95}                 & 2021          & \multicolumn{1}{l|}{87.2}          & 86.6        & \multicolumn{1}{l|}{97.2}          & 93.0        & \multicolumn{1}{l|}{94.4}          & 68.4        & \multicolumn{1}{l|}{68.1}          & 60.6        & \multicolumn{1}{l|}{76.7}          & 76.7        \\ \hline
HeteGCN (X-TX-X) \cite{c95}              & 2021          & \multicolumn{1}{l|}{86.3}          & 85.6        & \multicolumn{1}{l|}{97.3}          & 93.4        & \multicolumn{1}{l|}{93.3}          & 56.6        & \multicolumn{1}{l|}{66.7}          & 58.0        & \multicolumn{1}{l|}{77.6}          & 77.6        \\ \hline
HeteGCN (TX-X)   \cite{c95}              & 2021          & \multicolumn{1}{l|}{86.6}          & 86.0        & \multicolumn{1}{l|}{97.5}          & 93.9        & \multicolumn{1}{l|}{93.8}          & 65.3        & \multicolumn{1}{l|}{68.9}          & 61.8        & \multicolumn{1}{l|}{76.5}          & 76.5        \\ \hline
HGAT\tnote{6} \hspace{1sp} \cite{c131}         & 2021          & \multicolumn{1}{l|}{}              &             & \multicolumn{1}{l|}{}              &             & \multicolumn{1}{l|}{}              &             & \multicolumn{1}{l|}{42.7}          & 24.8        & \multicolumn{1}{l|}{62.8}          & 62.4        \\ \hline
MP-GCN \cite{c97}              & 2022          & \multicolumn{1}{l|}{86.8}          &             & \multicolumn{1}{l|}{97.8}          &             & \multicolumn{1}{l|}{94.5}          &             & \multicolumn{1}{l|}{70.3}          &             & \multicolumn{1}{l|}{77.9}          &             \\ \hline
ME-GCN\tnote{4} \hspace{1sp} \cite{c96}          & 2022          & \multicolumn{1}{l|}{28.6}          &             & \multicolumn{1}{l|}{86.8}          &             & \multicolumn{1}{l|}{78.3}          &             & \multicolumn{1}{l|}{27.4}          &             & \multicolumn{1}{l|}{68.1}          &             \\ \hline
ST-TextGCN\tnote{5} \hspace{1sp} \cite{c98}     & 2022          & \multicolumn{1}{l|}{}              &             & \multicolumn{1}{l|}{}              &             & \multicolumn{1}{l|}{}              &             & \multicolumn{1}{l|}{42.4}          & 25.1        & \multicolumn{1}{l|}{72.4}          & 72.4        \\ \hline
NNC-GCN \cite{xiao2024nnc}      & 2024         & 
\multicolumn{1}{l|}{}              &             & \multicolumn{1}{l|}{97.6}              &             & \multicolumn{1}{l|}{94.3}              &             & \multicolumn{1}{l|}{}          &        & \multicolumn{1}{l|}{}          &        \\ \hline
\end{tabular}
\label{tab6}
\begin{tablenotes}
    \item [1] 20 labeled data per class as reported in \cite{c24}
    \item [2] 25\% of classes unseen as reported in \cite{c94}
    \item [3] 20\% stratified sample of training documents as in \cite{c95}
    \item [4] 1-99 train/test split as reported in \cite{c96}
    \item [5] 10-90 train/test split as reported in \cite{c98}
    \item [6] 40 labeled data per class as reported in \cite{c131, c99}
\end{tablenotes}
\end{threeparttable}
}
\end{table*}

\begin{table*}[b]
\centering
\resizebox{0.8\textwidth}{!}{
\begin{threeparttable}[b]
\caption{\small Test accuracies and macro-averaged F1 scores for various self-supervised approaches under different conditions.}
\begin{tabular}{|l|l|l|ll|ll|ll|ll|l|l|}
\hline
\textbf{Method}              & \textbf{Year} 
& \textbf{20 NG}               
& \multicolumn{2}{l|}{\textbf{R8}}                 
& \multicolumn{2}{l|}{\textbf{R52}}                
& \multicolumn{2}{l|}{\textbf{Ohsumed}}            
& \multicolumn{2}{l|}{\textbf{MR}}                 
& \textbf{IMDB}              
& \textbf{Yelp}               \\ \hline
\textbf{}                    
& \textbf{}     
& \textbf{Acc.}
& \multicolumn{1}{l|}{\textbf{Acc.}} 
& \textbf{F1} 
& \multicolumn{1}{l|}{\textbf{Acc.}} 
& \textbf{F1} 
& \multicolumn{1}{l|}{\textbf{Acc.}} 
& \textbf{F1} 
& \multicolumn{1}{l|}{\textbf{Acc.}} 
& \textbf{F1} 
& \textbf{Acc.}
& \textbf{Acc.}  \\ \hline
GCNW-FL \cite{c101}          
& 2021          
& \textbf{}                      
& \multicolumn{1}{l|}{}              
&             
& \multicolumn{1}{l|}{}              
&             
& \multicolumn{1}{l|}{}              
&             
& \multicolumn{1}{l|}{}              
&             
& 68.5                 
& 62.7                      \\ \hline
CGA2-TC \cite{c99}          
& 2022          
& \textbf{}                       
& \multicolumn{1}{l|}{97.8}          
& 94.2        
& \multicolumn{1}{l|}{94.5}          
& 73.3        
& \multicolumn{1}{l|}{70.6}          
& 65.0        
& \multicolumn{1}{l|}{77.8}          
& 77.3                
&                           
&             \\ \hline
CGA2-TC\tnote{6} \hspace{1sp} \cite{c99}              
& 2022          
& \textbf{}                         
& \multicolumn{1}{l|}{}              
&             
& \multicolumn{1}{l|}{}              
&             
& \multicolumn{1}{l|}{50.7}          
& 41.9        
& \multicolumn{1}{l|}{52.0}          
& 51.8                    
&                          
&             \\ \hline
Cont-GCN-BERT \cite{c100}    
& 2023          
& 89.4                     
& \multicolumn{1}{l|}{98.3}          
&             
& \multicolumn{1}{l|}{96.9}          
&             
& \multicolumn{1}{l|}{73.1}          
&             
& \multicolumn{1}{l|}{86.4}          
&                          
&                           
&             \\ \hline
Cont-GCN-XLNet \cite{c100}   
& 2023          
& 89.7                     
& \multicolumn{1}{l|}{98.5}          
&             
& \multicolumn{1}{l|}{97.0}          
&             
& \multicolumn{1}{l|}{73.1}          
&             
& \multicolumn{1}{l|}{88.7}          
&                           
&                           
&             \\ \hline
Cont-GCN-RoBERTa \cite{c100} 
& 2023          
& 90.1                   
& \multicolumn{1}{l|}{98.6}          
&             
& \multicolumn{1}{l|}{96.6}          
&             
& \multicolumn{1}{l|}{73.4}          
&             
& \multicolumn{1}{l|}{91.3}         
&             
&                         
&                    \\ \hline
\end{tabular}
\label{tab7}
\begin{tablenotes}
    \item [6] 40 labeled data per class as reported in \cite{c131, c99}
\end{tablenotes}
\end{threeparttable}
}
\end{table*}

For text classification, Deferrard et al. \cite{c15} applied their approach to the 20 NG dataset. While it was only able to attain second place next to the multinomial Naive Bayes classifier with an accuracy of 68.3, it outperformed traditional fully connected networks while having fewer parameters. This approach was later formalized and further generalized for improved scalability and classification performance in large-scale networks by \cite{c16}. \cite{c17} focused exclusively on text classification and proposed a GCN based on \cite{c16} and formulated using a novel heterogeneous graph approach. Experiments on 20 NG, R8, R52, MR, and Ohsumed demonstrated that their proposed method was able to outperform existing state-of-the-art solutions without relying on any pre-trained word embeddings, especially when training data was scarce. This would provide the basis and serve as a frequent benchmark for all subsequent approaches. Works that immediately followed, such as \cite{c18} and \cite{c19}, sought to optimize the approach in \cite{c17}, and while there was only an incremental improvement in classification performance, it was clear that proposing more computationally efficient, scalable, and robust models was their primary goal. Later, \cite{c20} additionally optimized this approach for handling over-smoothing due to a higher number of graph convolutions using the Markov Diffusion Kernel, while more recently, \cite{c129} opted for a different route by addressing local intra-class diversity and local inter-class similarity that are implicitly encoded within the graph structure. In general, these approaches improved TextGCN performance in different ways without relying on pre-trained embeddings or any extrinsic knowledge sources. These are, however, primarily transductive in nature.  In terms of test accuracy, the most notable improvements in classification performance came when researchers attempted to rework \cite{c17} by either enriching the graph representation to capture more textual context \cite{c23} or by augmenting the GCN with other models/embeddings such as BERT \cite{c21}, \cite{c22} to achieve a solution that was, in theory, greater than the sum of its parts. Among these divergent approaches, the latter yielded the most promising results. In this regard, \cite{c22} has generally attained SOTA performance across multiple datasets (Table \ref{tab5}).

Another line of research attempted to extend the existing transductive GCN-based approaches to inductive settings to allow for online testing, that is to generalize patterns and relationships from the training data to make accurate predictions on new, unseen instances \cite{c25}, \cite{c26}. These methods were able to demonstrably outperform TextGCN under inductive constraints such as limited labeled data. InducT-GCN, in particular, bested various GCN-based baselines and certain models using pre-trained embeddings owing to its better generalization capabilities.
Similar to inductive approaches, semi-supervised approaches also demonstrate their efficacy with limited labeled datasets as they are able to compensate for this constraint by extracting and supplementing additional information from unlabeled data and improve model performance. In literature, researchers have employed a variety of subsetting and sampling techniques to demonstrate the effectiveness of their proposed models in limited labeled settings. While this makes it difficult to draw a definitive conclusion about the overall state of semi-supervised approaches, we can still gain valuable insights by considering common baselines.

As graph convolutions enable information sharing among neighboring nodes through the inherent message propagation mechanism, GCN models can transfer knowledge from labeled nodes to unlabeled ones and essentially leverage the underlying graph structure to improve their ability to make predictions on the unlabeled data. While this was successfully demonstrated early on for citation networks in \cite{c16}, semi-supervised GCN approaches for modeling free text are fairly recent. The authors of \cite{c24} claimed their approach to be the first to model free text under strict semi-supervised conditions and demonstrated their model’s performance gain over \cite{c17} and \cite{c23} when using 20 labeled samples per class for training. Different variations of \cite{c95} were also able to outperform \cite{c17} across the board with regards to test accuracy and f1 scores, albeit under different limited labeled data conditions (20\% stratified sample of training documents as in \cite{c17}). However, the performance gains were not as pronounced in the large labeled data scenario. The same was also true for the approach in \cite{c97}, which reported slight gains over \cite{c95} under the same large labeled conditions. \cite{c131}, \cite{c96}, and \cite{c98} also exhibited improved performance over \cite{c17} across multiple datasets under their respective limited labeled conditions (Table \ref{tab6}).
Self-supervised GCN techniques for text classification have also begun to gain prominence in recent years. Like semi-supervised approaches, these are also able to leverage unlabeled data to improve model performance. \cite{c99} was able to attain performance comparable to the best supervised approach that did not use pre-trained embeddings in the large labeled setting and that of \cite{c131} in the same limited labeled setting. Moreover, \cite{c100} utilized various pre-trained language models along with their proposed ATAD scheme and reportedly outperformed all aforementioned methods in offline settings while also faring similarly well in online settings (Table \ref{tab7}). However, it should be noted that this approach is very recent and requires further validation and scrutiny by the research community at large. Nonetheless, it offers valuable insight into the trajectory of research in this domain and could be a promising avenue for future exploration.

\section{Conclusion and Future Research Directions}
Graph Convolutional Networks hold great promise for addressing text classification, as they have demonstrated impressive results in various studies and benchmarks. However, there are still many challenges and research directions to explore in order to improve their effectiveness and efficiency in this domain.

{Deep graph learning with limited-labeled data or noisy data can hinder the performance of GCNs and their generalization in real-world scenarios. Conventional data augmentation techniques often fall short in addressing data scarcity and noise in graph structures. Researchers are exploring specialized augmentation methods, such as transforming the graph adjacency matrix or the node feature matrix, or using label enrichment \cite{c132}. However, most existing approaches rely on handcrafted strategies based on performance on downstream tasks like text classification or recommendation \cite{c133}, limiting their practical use without abundant labeled data. Developing dynamic data augmentation algorithms that automatically apply optimal transformations and perturbations is crucial. Additionally, preserving the semantics encoded within the graph structure is essential. In text graphs, nodes represent documents and words, while edges represent relationships based on criteria like word co-occurrence, TF-IDF, or semantic similarity. Effective augmentation strategies should maintain these semantics, potentially guided by pre-trained word embeddings (e.g., BERT, GPT), ontologies like WordNet, and syntactic and semantic rules. Graph diffusion algorithms can also facilitate data augmentation by iteratively updating node features based on neighbors, exploiting the global structure knowledge of the graph. This approach generates augmented samples with updated node features, enriching the dataset and improving the model's robustness to variations in information propagation, enhancing its understanding of the graph's structure.}

{Like graph convolution, graph diffusion also utilizes the graph structure to model information propagation and relationships between nodes. However, unlike the former which captures local patterns and relationships by aggregating information from a node’s immediate neighbors, graph diffusion focuses on integrating global influences and properties across the entire graph. Such local and global information integration is crucial for understanding the nuances of language and particularly beneficial in text classification tasks, as demonstrated by combining GCN with BERT models \cite{c21, c22}. Recent studies suggest implementing diffusion either as a preprocessing step to enhance the initial graph structure \cite{c134} or as an adaptation within the GCN architecture \cite{c135}, promoting a more robust and context-aware learning process. These works report improvements in node classification, underscoring the potential for improved performance on complex text classification tasks.}

{For instance, improved data augmentation and diffusion techniques could significantly enhance performance in sentiment analysis, topic detection, and document classification. By effectively handling noisy and limited-labeled data, GCNs can be more robust in real-world applications, such as in large-scale social media monitoring, customer feedback analysis, and automated content moderation.}

{Integrating GCNs with LLMs like GPT is another promising direction. GCNs capture structural information in text data, while LLMs excel in understanding and generating human-like text. Combining these strengths can help create more robust, context-aware models. GCNs can enhance LLMs by embedding word and sentence relationships into the learning process, improving text classification, sentiment analysis, and other NLP tasks. This approach is especially effective in domains needing a deep understanding of semantic content and contextual relationships, such as legal document analysis, biomedical text mining, and social media analytics.}

{In the context of privacy protection, GCNs offer unique advantages. As data privacy becomes increasingly important, especially with regulations like GDPR and CCPA, it is crucial to develop models that can operate efficiently while ensuring user data is protected. \cite{liang2024graph} employed GCNs in a novel model that integrates text data and label correlations, utilizing a double-attention mechanism to significantly enhance detection performance for privacy disclosures in online posts. By leveraging GCNs' ability to understand complex relationships among different types of private information, we can improve privacy detection tools for social media and similar platforms, effectively mitigating potential risks. This also makes GCNs suitable for applications in healthcare, finance, and other sectors handling sensitive information.
}

\nocite{*}


\end{document}